\documentclass[pdflatex, sn-mathphys-num]{sn-jnl}

\usepackage{graphicx}%
\usepackage{multirow}%
\usepackage{amsmath,amssymb,amsfonts}%
\usepackage{amsthm}%
\usepackage{mathrsfs}%
\usepackage[title]{appendix}%
\usepackage{xcolor}%
\usepackage{textcomp}%
\usepackage{manyfoot}%
\usepackage{booktabs}%
\usepackage{algorithm}%
\usepackage{algorithmicx}%
\usepackage{algpseudocode}%
\usepackage{listings}%
\usepackage{bm}
\usepackage{natbib}
\newgeometry{margin=1in}  

\usepackage{anyfontsize}  
\usepackage{cleveref}  
\usepackage{soul, color}  
\usepackage{mdframed}  
\usepackage{pdflscape}  
\usepackage[labelformat=empty,skip=0pt]{subcaption} 

\hypersetup{
    pdfauthor={Joshua E. Hammond, Tyler Soderstrom, Brian A. Korgel, Michael Baldea},
    pdftitle={Staying Alive: Online Neural Network Maintenance and Systemic Drift},
}

\begin{document}
\renewcommand{\cal}{\mathcal}
\newcommand{\beqn}{\begin{eqnarray}}
        \newcommand{\bdm}{\begin{displaymath}}
                \newcommand{\edm}{\end{displaymath}}
        \newcommand{\eeqn}{\end{eqnarray}}
\newcommand{\be}{\begin{equation}}
        \newcommand{\ee}{\end{equation}}
\newcommand{\ba}{\begin{array}}
        \newcommand{\ea}{\end{array}}
\newcommand{\pa}{\partial}
\newcommand{\re}{\ref}
\newcommand{\ci}{\cite}
\newcommand{\la}{\label}
\newcommand{\lr}{\leftrightarrow}
\newcommand{\fr}{\frac}
\newcommand{\ov}{\overline}
\newcommand{\Om}{\Omega}
\newcommand{\lb}{\lambda}
\newcommand{\ga}{\gamma}
\newcommand{\te}{\theta}
\newcommand{\na}{\nabla}
\newcommand{\om}{\omega}
\newcommand{\bt}{\beta}
\newcommand{\al}{\alpha}
\newcommand{\si}{\sigma}
\newcommand{\bmzeta}{{\bm {\zeta}}}
\newcommand{\bmeta}{{\bm {\eta}}}
\newcommand{\ovy}{\overline{Y}}
\newcommand{\ar}{\leftrightarrow}
\newcommand{\ri}{\rightarrow}
\newcommand{\Lr}{\Longrightarrow}
\newcommand{\tocEF}{\st{{\cal E}_F}\lr}
\newcommand{\tocM}{\st{{\cal M}}\lr}
\newcommand{\tocF}{\st{{\cal F}}\lr}
\newcommand{\toM}{\st{M}\lr}
\newcommand{\ffi}{\varphi}
\newcommand{\eps}{\varepsilon}
\newcommand{\de}{\delta}
\newcommand{\ds}{\displaystyle}
\newcommand{\De}{\Delta}
\newcommand{\ti}{\tilde}
\newcommand{\percent}{\%}
\newcommand{\rhu}{\rightharpoonup}
\newcommand{\und}{\underline}
\newcommand{\non}{\nonumber}
\newcommand{\dis}{\displaystyle}
\newcommand{\deh} {\Delta H}
\newcommand{\etz} {\eps \rightarrow 0}
\newcommand{\oeps} {\cal{O}(\eps)}
\newcommand{\oone} {{\cal{O}}(1)}
\newcommand{\ooveo} {{\cal{O}}(\frac{1}{\eps})}
\newcommand{\oove} {\frac{1}{\eps}}
\newcommand{\ooveone} {\frac{1}{\eps_1}}
\newcommand{\ooved}{\frac{1}{\eps_2}}
\newcommand{\ooveu} {\frac{1}{\eps_1}}
\newcommand{\doveu} {\frac{\eps_2}{\eps_1}}
\newcommand{\barr}{\ba}
\newcommand{\earr}{\ea}
\newcommand{\lbr}{\left\lbrace}
\newcommand{\rbr}{\right\rbrace}
\newcommand{\uT}{\und{T}}
\newcommand{\ux}{\und{x}}
\newcommand{\uX}{\und{X}}
\newcommand{\ug}{\und{g}}
\newcommand{\uf}{\und{f}}
\newcommand{\cpl}{C_{p,l}}
\newcommand{\cpv}{C_{p,v}}
\newcommand{\cR}{\cal R}
\newcommand{\bfx}{{\bf x}}
\newcommand{\bfl}{{\bf l}}
\newcommand{\bfc}{{\bf c}}
\newcommand{\bfy}{{\bf y}}
\newcommand{\bfg}{{\bf g}}
\newcommand{\bff}{{\bf f}}
\newcommand{\bfu}{{\bf u}}
\newcommand{\bfz}{{\bf z}}
\newcommand{\bfk}{{\bf k}}
\newcommand{\bfh}{{\bf h}}
\newcommand{\btf}{{\bf \tilde f}}
\newcommand{\bbf}{{\bf \bar f}}
\newcommand{\ka}{\kappa}
\newcommand{\bfG}{{\bf G}}
\newcommand{\bfp}{{\bf p}}
\newcommand{\bfr}{{\bf r}}
\newcommand{\bfF}{{\bf F}}
\newcommand{\bfB}{{\bf B}}
\newcommand{\bfL}{{\bf L}}
\newcommand{\bfe}{{\bf e}}
\newcommand{\bfC}{{\bf C}}
\newcommand{\bfo}{{\bf 0}}
\newcommand{\bmth}{{\bm \theta}}
\newcommand{\bmr}{{\bm \rho}}
\newcommand{\bmz}{{\bm \zeta}}
\newcommand{\bmg}{{\bm \gamma}}
\newcommand{\bmG}{{\bm \Gamma}}
\newcommand{\bmp}{{\bm \pi}}
\newcommand{\bfS}{{\bf S}}
\newcommand{\bmf}{{\bm \phi}}
\newcommand{\bmj}{{\bm j}}
\def\R{{\rm I\kern-.19em R}}
\def\M{{\rm I\kern-.1567em M}}
\def\C{{\rm I\kern-.55em C}}
\def\intp{{\rm \int{\hspace{-4mm}}-}}
\def\tr{{\rm tr}}
\def\Id {{\rm Id}}
\def\div {{\rm div}}
\def\sign {{\rm sign}}
\def\spt{{\rm spt}}
\def\dist{{\rm dist}}
\def\eg{{\it e.g.,}~}
\def\ie{{\it i.e.,}~}
\def\Chi{{\raise.4ex\hbox{\large$\chi$}}}
\def\Z{\cal{Z}}
\newtheorem{theorem}{Theorem}

\newtheorem{claim}[theorem]{Claim}
\newtheorem{example}[theorem]{Example}


\newcommand{\jac}[1]{\nabla_{#1} \bff}
\newcommand{\gradloss}{\nabla_{\bmp} \mathcal{L}}
\newcommand{\agradloss}{|\nabla_{\bmp} \mathcal{L}|}
\newcommand{\TODO}[1]{\hl{TODO: #1}}
\newcommand{\needscitation}{\hl{[citation needed]}}
\newcommand{\subitemize}[1]{\begin{itemize}\item #1\end{itemize}}  
\newcommand*\phantomsubfigure[1]{\begin{subfigure}[]{0pt}\caption{}\label{#1}\end{subfigure}}  

\title{Staying Alive: Online Neural Network Maintenance under Systemic Drift}






\author[1]{\fnm{Joshua E.} \sur{Hammond}}\email{joshua.hammond@utexas.edu}

\author[2]{\fnm{Tyler} \sur{Soderstrom}}\email{tyler.a.soderstrom@exxonmobil.com}

\author[1,3]{\fnm{Brian A.} \sur{Korgel}}\email{korgel@che.utexas.edu}

\author*[1,4]{\fnm{Michael} \sur{Baldea}}\email{mbaldea@che.utexas.edu}

\affil[1]{\orgdiv{McKetta Department of Chemical Engineering}, \orgname{The University of Texas at Austin}, \orgaddress{\street{200 E. Dean Keeton St.\ Stop C0400}, \city{Austin}, \postcode{78712}, \state{Texas}, \country{USA}}}

\affil[2]{\orgname{ExxonMobil Technology and Engineering}, \city{Spring},\state{Texas}, \country{USA}}

\affil[3]{\orgdiv{Energy Institute}, \orgname{The University of Texas at Austin}, \orgaddress{\street{2304 Whitis Ave.\ Stop C2400}, \city{Austin}, \postcode{78712}, \state{Texas}, \country{USA}}}

\affil[4]{\orgdiv{Institute for Computational Engineering and Sciences}, \orgname{The University of Texas at Austin}, \orgaddress{\street{201 E.\ 24th Street, POB 4.102, Stop C0200}, \city{Austin}, \postcode{78712}, \state{Texas}, \country{USA}}}


\abstract{We present the Subset Extended Kalman Filter (SEKF) as a method to \emph{update} previously trained model weights online rather than retraining or finetuning them when the system a model represents drifts away from the conditions under which it was trained. We identify the parameters to be updated using the gradient of the loss function and use the SEKF to update only these parameters. We compare finetuning and SEKF for online model maintenance in the presence of systemic drift through four dynamic regression case studies and find that the SEKF is able to maintain model accuracy as-well if not better than finetuning while requiring significantly less time per iteration, and less hyperparameter tuning.}




\maketitle


Deep learning is an increasingly widespread method for mapping inputs to outputs in complex conditions where first-principles methods are impractical due to temporal or computational constraints~\cite{zeng2019continual, santander2023deep, kumar2023grey, lejarza2022data}. Currently, deep learning models are developed in an offline train-then-deploy paradigm using past data; However, this does not reflect many realistic applications where the underlying data-generating physical systems are constantly evolving. Dynamical systems naturally change and may drift away from the conditions under which the model was trained, causing prediction or classification error to increase\cite{bayram2022concept, ackerman2021automatically, pears2014detecting}. In these cases, the model must be altered or revised to ensure that it remains accurate and reliable. We will refer to this updating process as model maintenance.

The principal criterion for successful model maintenance is accuracy in performing the intended task (such as prediction or classification). Other desirable features include computational resource efficiency, workflow sustainability, use of previous knowledge to accelerate assimilation of new data, automatic, autonomous learning without manual supervision or tuning, and robustness to noise~\cite{verwimp2023continual,veniat2020efficient,van2024continual,gama2014survey,jiang2024network}

In practice, incremental, offline model maintenance usually occurs when a drop in model performance is detected. The two most common approaches include \emph{stateless} model retraining which re-initializes and retrains all model parameters, and \emph{stateful} training or updates such as finetuning where model parameters are modified beginning with their current values\cite{huyen2022designing}\footnote{note that it is not ``stateful retraining'' because there is no \emph{re-} training here, only continual adaptation}. Each of these carries their own risks: retraining is computationally expensive and may require a large amount of data, time, and computational expense, while incremental finetuning may lead loss of generalization, and provides a larger infrastructure challenge\cite{Ash2019,berariu2021study}. Both methods involve tuning hyperparameters (when updates should occur, what data to include when training, learning rates, etc.).

Online learning is an active area of research that aims to update models using a stream of new data (that are not necessarily \textit{iid}) as they becomes available. Online learning offers a promising alternative to inefficient stateless model retraining\cite{jiang2024network} and has already been shown to require lower computational cost and achieve better performance than stateless offline approaches\cite{Egg2021}.

We are interested in developing an online method to maintain previously trained neural network models of dynamical systems that obey material and energy conservation laws for applications such as control and optimization. These applications require a high degree of model accuracy and fast execution of model prediction, as well as fast execution of model maintenance operations.

Taking inspiration from singular perturbation theory, we will hypothesize that in the presence of gradual, monotonic drift that does not change the overall system structure, it may be possible to identify and update only a small subset of model parameters to maintain model accuracy. To test this hypothesis, we formulate and address two key questions:

\begin{itemize}
    \item \emph{Which} parameters should be updated?
    \item \emph{How} should parameters be updated?
\end{itemize}

In the context of neural network models, we show that using the gradient of loss is an effective means to identify the parameters with the largest contribution to prediction error. To the second point, introduce the Subset Extended Kalman Filter (SEKF) to update the selected parameters at minimal computational cost. We demonstrate the effectiveness of the SEKF to update model parameters in the presence of drift and compare the results to offline retraining and finetuning methods using synthetic examples as well as case studies from chemical industry and healthcare.

\section*{Nomenclature}
\begin{table}[h]
    \caption{Symbols and Abbreviations}
    \label{tab:nomenclature}
    \begin{tabular}{ll}
        \toprule
        Symbol                           & Description                                                                                                                                  \\ \midrule
        $\jac{\bfp}$                     & Jacobian matrix of $\bff$ with respect to $\bfp$                                                                                             \\
        $\gradloss$                      & Gradient of loss function with respect to model parameters $\bmp$                                                                            \\
        $\epsilon$                       & Scaling parameter ($0<\epsilon\ll 1$)                                                                                                        \\
        $\eta_k$                         & Learning rate at sample time $k$                                                                                                             \\
        $\bmp$                           & Parameters of Neural Network ($\mathcal{D}^{\pi}\subset \R^l$)                                                                               \\
        $\phi(\hat \bfx,\hat \bfu,\bmp)$ & ANN model                                                                                                                                    \\
        $\tau$                           & Slow timescale ($\tau = \epsilon t$)                                                                                                         \\
        ANN                              & Artificial Neural Network                                                                                                                    \\
        $\mathcal{D}^i$                  & Domain of variable $i$                                                                                                                       \\
        $\bfe$                           & Error                                                                                                                                        \\
        $\bff(\bfx,\bfu,\bfp)$           & ODE system                                                                                                                                   \\
        $\bfg(\bfx,\bfp)$                & Parameter Drift Model                                                                                                                        \\
        NODE                             & Neural Ordinary Differential Equation                                                                                                        \\
        $K$                              & Span of training data                                                                                                                        \\
        $k$                              & Sample time                                                                                                                                  \\
        $H_k$                            & Flattened jacobian $\nabla_{\pi_k}\hat{\bfx}_k$                                                                                              \\
        $\bmj$                              & Index of selected parameters                                                                                                                 \\
        $\mathcal{L}$                    & Loss                                                                                                                                         \\
        Mag.$q$                          & Parameter selection function all parameters with $\agradloss$ larger than $\text{quantile}(|\nabla_{\pi 0}\mathcal{L}_\text{validation}|,q)$ \\
        MTPI                             & Mean Time Per Iteration (s)                                                                                                                  \\
        $\bfp$                           & Parameters of ODE System $\bff$ ($\mathcal{D}^p\subset \R^k$)                                                                                \\
        Prop.$q$                         & Parameter selection function updating the highest $q$ quantile of parameters based on $\agradloss$                                           \\
        $P_k$                            & Covariance matrix of model variables at sample time $k$                                                                                      \\
        $Q_k$                            & Process noise covariance matrix at sample time $k$                                                                                           \\
        $R_k$                            & Modified measurement noise covariance matrix at sample time $k$                                                                              \\
        $T_s$                            & Sample time                                                                                                                                  \\
        $t_s$                            & Update interval                                                                                                                              \\
        $\bfu$                           & Known inputs ($\mathcal{D}^u\subset \R^m$)                                                                                                   \\
        $\hat \bfu_k$                    & Measured inputs at sample time $k$                                                                                                           \\
        $w$                              & Gaussian noise ($w\sim \cal N(0,\sigma)$)                                                                                                    \\
        $\bfx$                           & State variables ($\mathcal{D}^x\subset \R^n$)                                                                                                \\
        $\hat \bfx_k$                    & Measured state variables at sample time $k$                                                                                                  \\
        $\tilde \bfx_{k+1}$              & Output of a model, Prediction of state variables at sample time $k+1$                                                                        \\
        \bottomrule
    \end{tabular}
\end{table}

\section{Background}\label{sec:background}
We are concerned with Artificial Neural Network (ANN) models for physical systems that obey conservation laws for mass and energy and are governed by ordinary differential equation systems of the form:
\begin{equation}
    \label{eq:governingequation}
    \frac{d \bfx}{dt} = \bff(\bfx,\bfu,\bfp)
\end{equation}
with state variables $\bfx \in \cal D^x \subset \R^n$, known inputs $\bfu \in \cal D^u \subset \R^m$, parameters $\bfp \in \cal D^p \subset \R^k$. The function $\bff$ is such that $\bff: \cal D^x \times \cal D^u \times \cal D^p \rightarrow \cal D^x$.
Models of this type are representative of the dynamic behavior of industrial \cite{venkatesan2024recent,  kumpati1990identification, park2023simultaneous, luo2023model, yu2002model,zorriassatine1998review}, and biological systems \cite{elmarakeby2021biologically, psichogios1992hybrid, titano2018automated}, and find applications for example in real-time control\cite{emami2022neural,karabacak2014artificial}, fault detection\cite{li2022defect, hoskins1988artificial, watanabe1989incipient}, etc.

We will consider the circumstance where physical systems represented by models such as \eqref{eq:governingequation} are subject to parametric drift. We define drift as a change in the parameters $\bfp$ that occurs over a time scale $\tau$ that is much longer/slower than the time scale of the governing equations, that is,
\beqn
\label{eq:pdynamics}
\frac{d \bfp}{d\tau} = \bfg(\bfx,\bfp) \qquad \bfg: \cal D^x \times \cal D^p \rightarrow \cal D^p
\eeqn
where
\be
\tau = \epsilon t\qquad 0< \epsilon \ll 1
\ee
The evolution in time of the parameter vector $\bfp$ alters the state variables $\bfx$ even when the inputs $\bfu$ do not vary. If the Jacobian matrix $\jac{}$ is nonsingular at any point in $\cal D^x \times \cal D^u \times \cal D^p$, parameter drift does not lead to any structural changes in the function $\bff$ or alter the topological behavior of the system \eqref{eq:governingequation}.

We assume that these models are either unknown, too complicated, or too computationally intensive to produce results in a practically meaningful amount of time. Instead, a sufficiently large set of measurement data (corrupted with noise $w$, assumed to be Gaussian, $w \sim \cal N(0,\sigma)$) is presented in the form of input/output pairs $(\hat \bfu_k, \hat \bfx_k)$ at sample times $k\in \{0,\dots,K\}$ are available, with sample time $t_s$.

According to the universal approximation theorem\cite{hornik1989multilayer}, these data can be used to construct and train an ANN model of the system \eqref{eq:governingequation}, of the form:

\be
\label{eq:ann1}
\tilde \bfx_{k+1} = \phi(\hat \bfx_k, \hat \bfx_{k-1}, \dots, \hat \bfu_{k}, \hat \bfu_{k-1}, \dots, \bmp)
\ee
where $\tilde \bfx_{k+1}$ represents the \textit{prediction} of the values of the state variables at the sample time $k+1$, made on the basis of the \textit{measured} values of the states and inputs at sample time $k$, and possibly a finite number of previous sample times $k-1, \dots$ and $\bmp \in \cal D^{\pi} \subset \R^l$ the parameters of the ANN. For brevity, we will refer to $\phi(\hat \bfx_k, \hat \bfx_{k-1}, \dots, \hat \bfu_{k}, \hat \bfu_{k-1}, \dots,\bmp)$ as $\phi(\cdot,\cdot,\bmp)$. We note that the dimension $l$ of vector $\bmp$ can be quite large, particularly in the case of ``deep'' network architectures featuring many layers, and also point out that $\bmp$ are neither the same nor have the same physical meaning as $\bfp$.

Such ANN models aim to capture the governing dynamics of the system (Equation \ref{eq:governingequation}) and often remain agnostic to the parameter drift \eqref{eq:pdynamics}. There are two related causes for this; first, for practical reasons, the time span $K$ of the data available for training the ANN may be short compared to the scale $\tau$ of the evolution of the parameters $\bfp$. Second, that -- in the time span covered by the data --  the changes to $\bfx$ that are due to the drift of $\bfp$ may not be detectable over the impact of measurement noise (equivalently, that signal to noise ratio is too low to discern the impact of drift). It is also worth mentioning here that parameters $\bfp$ are often not measured directly and hence explicitly capturing their effect on the system behavior (and reflecting this effect in the predictions of the ANN) remains difficult. Consequently, parameter drift is expected to cause the predictions $\tilde \bfx$ of the model \eqref{eq:ann1} to gradually diverge from the corresponding values $\hat \bfx$ that are measured from the evolution of the underlying dynamical system \eqref{eq:governingequation}. Equivalently, the error
\beqn
\bfe_k = \hat \bfx_k - \tilde \bfx_k
\label{eq:error}
\eeqn
tends to increase (in a norm sense) over time.

Motivated by the above, this work explores new means of correcting this discrepancy. Frequently, this is addressed by developing a \textit{closure}, i.e., a model of the error $\bfe$ that can then be used to correct the model output $\tilde \bfx$. Here, we propose a different tack. Specifically, we hypothesize that a small subset of the parameters $\bmp$ of the neural network can be altered to correct for drift, and demonstrate a mechanism based on Kalman filtering for doing so.

\section{Methodology}\label{sec:methodology}

\subsection{Singular Pertubation Analysis of Multi-Scale Systems}
Together, the governing equations \eqref{eq:governingequation} and parametric drift model \eqref{eq:pdynamics} form an ordinary differential equation system in the standard singularly perturbed form\cite{Fenichel1979}:
\begin{subequations}
    \label{eq:goveq2ts}
    \beqn
    \frac{d \bfx}{dt} &=& \bff(\bfx,\bfu,\bfp)\\
    \frac{d \bfp}{dt} &=& \eps \bfg(\bfx,\bfp)
    \eeqn
\end{subequations}
where $\bfx$ are the fast states, and $\bfp$ are the slow states, $t$ represents the fast time scale, and $\tau$ represents the slow time scale.

To build our argument, we will first consider the case where the governing equations shown in \eqref{eq:goveq2ts} as well as all states $\bfx$ and parameters $\bfp$ are available and will be used for predicting the future state of the system $\bfx_{k+1}$ based on the current state $\bfx_k$ and input values $\bfu_k$. As pointed out earlier, in practice, it is not typical to measure $\bfp$ in real time, nor is there an interest in predicting its values. Rather, it is important that the states $\bfx$ be predicted accurately, that is, the output of any model, $\tilde \bfx$, should be close in a norm sense to the measured data $\hat \bfx$.

The analysis of systems of the type \eqref{eq:goveq2ts} can be carried out using the geometric singular perturbation theory of  \citet{Fenichel1979}. It is typical to seek an approximation of the \textit{slow} dynamics. To this end, an asymptotic approximation is obtained in the limit case $\etz$ in the slow time scale. Expressing \eqref{eq:goveq2ts} in the time scale $\tau$, we obtain:
\begin{subequations}
    \label{eq:goveq2ts_1}
    \beqn
    \frac{d \bfx}{d\tau} &=& \oove \bff(\bfx,\bfu,\bfp)\\
    \frac{d \bfp}{d\tau} &=&  \bfg(\bfx,\bfp)
    \eeqn
\end{subequations}
then, in the limit $\etz$, an expression for the system dynamics in the slow time scale is obtained in the form of a constraint differential equation:
\begin{subequations}
    \label{eq:goveq2ts_2}
    \beqn
    \bfo &=& \bff(\bfx,\bfu,\bfp)\\
    \frac{d \bfp}{d\tau} &=&  \bfg(\bfx,\bfp)
    \eeqn
\end{subequations}
where the algebraic constraint $\bfo = \bff(\bfx,\bfu,\bfp)$ describes a slow manifold for the system dynamics. The assumption that $\nabla \bff$ is nonsingular ensures that the dynamics of the system in the slow time scale are well approximated by \eqref{eq:goveq2ts_2}.

While we present this result for the sake of completeness, it is not of interest to us, since our intent is to model/predict the future evolution of the \textit{fast} states $\bfx$ rather than that of the \textit{slow} states $\bfp$. A similar asymptotic analysis can be performed by considering the limit $\etz$ in the fast time scale $t$. In this case, we obtain the layer equation:

\begin{subequations}
    \label{eq:goveq_layereq}
    \beqn
    \frac{d \bfx}{dt} &=& \bff(\bfx,\bfu,\bfp)\\
    \frac{d \bfp}{dt} &=& \bfo
    \eeqn
\end{subequations}
The layer equation is an accurate representation of the fast dynamics (the evolution of $\bfx$) in ``layers'' where $\bfp$ does not change, and can be a good approximation of the fast dynamics as long as $\bfp$ has not changed much.

The evolution of $\bfx$ at small times $t$ can be captured by setting $\bfp=\bfp_o$, the initial condition of the slow states $\bfp$. The question is then what happens over longer time horizons, where the change in $\bfp$ becomes potentialy substantial. To address this, assume that values of $\bfp$ can be sampled at some sample time $T_s$ from the evolution of the system \eqref{eq:goveq2ts}, and that these values $\bfp_j$ are substituted periodically in \eqref{eq:goveq_layereq}, creating an iterative scheme where the system:
\begin{subequations}\label{eq:goveq_layereq_iterative}
    \beqn
    \frac{d \tilde \bfx}{dt} &=& \bff(\tilde \bfx,\bfu,\bfp_j)\qquad \forall t \in [t_j, t_j +T_s] \\
    \tilde \bfx(t_j) & = &\tilde \bfx(t_{j-1} + T_s)
    \eeqn
\end{subequations}
is integrated repeatedly over the time intervals $[t_j, t_j +T_s]$, during which $\bfp$ is held constant at the value $\bfp_j$  (i.e., sample and hold). We use the notation $\tilde \bfx$ to denote that the states of \eqref{eq:goveq_layereq_iterative} are to be considered model predictions in the sense introduced above. It is easy to show that in the limit of $T_s \rightarrow 0$, we have $\bfx \equiv \tilde \bfx$ and the iterative scheme \eqref{eq:goveq_layereq_iterative} is reduced to the model \eqref{eq:goveq2ts}.

We illustrate this scheme with a simple example.

\begin{example}\label{ex:simpleexample}
    We consider the singularly perturbed system

    \begin{subequations}\label{eq:example1}
        \beqn
        \frac{dp}{dt} &\triangleq& \eps (x - p) = \bfg(x,p)\\
        \frac{dx}{dt} &\triangleq& -px + 1 + u = \bff(x,u,p)
        \eeqn
    \end{subequations}
    which is of the form \eqref{eq:goveq2ts} with $\bfx = x$, $\bfp = p$. Let $x(0) = 0$, $p(0) = 1$, $\eps = 0.01$ $T_s=25$ and $u = -0.01 t$. The simulation results in \Cref{fig:boundarylayer_ode} show that the iterative scheme \eqref{eq:goveq_layereq_iterative} (values shown as the dashed black line) provides a vastly improved (albeit discontinuous) prediction of the values of the fast state $x$ relative to the layer equation \eqref{eq:goveq_layereq}.

    \begin{figure}[htb]
        \centerline{\includegraphics[width=183mm]{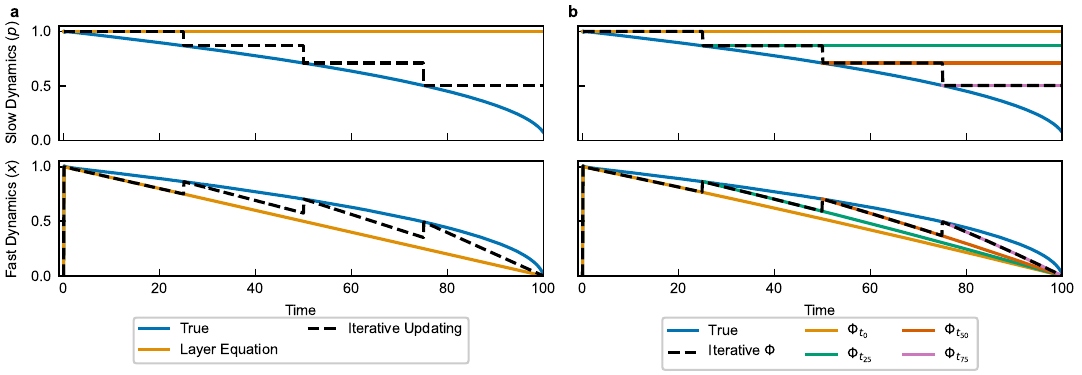}}
        \phantomsubfigure{fig:boundarylayer_ode}
        \phantomsubfigure{fig:boundarylayer_nn}
        \caption{Simulation of the system in Example~\ref{ex:simpleexample}. \textbf{a} Solution of the system \eqref{eq:example1} in blue, simulation of the layer equation~\eqref{eq:goveq_layereq} in orange, and the results of the iterative scheme~\eqref{eq:goveq_layereq_iterative} is shown with dashed lines. \textbf{b} a neural network approximation of the system where the neural network is iteratively retrained to reflect recent dynamics.}
    \end{figure}

\end{example}

\subsection{Updating ANN Models}
The findings above suggest that an iterative modeling scheme of the type \eqref{eq:goveq_layereq_iterative} can provide a good representation for the evolution of the fast variables $\bfx$, as long as a suitable way to update the model parameters $\bfp$ exists.The model \eqref{eq:goveq_layereq} is a phenomenological reduced-order model of the fast dynamics of the system \eqref{eq:goveq2ts}, that was derived from the governing equations themselves.

In light of the assumptions in Section \ref{sec:background}, the ANN model~\eqref{eq:ann1} is itself a representation of the fast dynamics of the system, defined by  $\phi(\cdot,\cdot,\bmp)$, with the parameter vector $\bmp$. Equivalently, \eqref{eq:ann1} is a functional approximator of \eqref{eq:goveq_layereq}.

While parameters $\bmp$ do not have the same physical meaning as $\bfp$, it is to be expected that they \emph{should} be evolving over time to reflect the evolution of parameters $\bfp$. We postulate that an iterative scheme of the type \eqref{eq:goveq_layereq_iterative}, when adapted to the ANN model, as shown below, can be used to improve the accuracy of the ANN model (reduce the error \eqref{eq:error}), assuming that a modality to update $\pi$ exists.
\begin{subequations}\label{eq:goveq_layereq_nn_iterative}
    \beqn
    \tilde \bfx_{k+1}& = &\phi(\hat \bfx_k, \hat \bfx_{k-1}, \dots, \hat \bfu_{k}, \hat \bfu_{k-1}, \dots, \bmp_j) \qquad \forall t \in [t_j, t_j +T_s] \\
    \tilde \bfx(t_j) & = &\tilde \bfx(t_{j-1} + T_s)
    \eeqn
\end{subequations}
Note that the updating interval $T_s \ge t_s$, i.e, updating parameters $\pi$, can happen at every sample time, but can also be done less frequently.

\begin{example}
    Consider an ANN with a simplified form of \eqref{eq:goveq_layereq_nn_iterative} shown in \eqref{eq:example1_nn}. $\bmp$ are trained such that they approximate the fast dynamics $\bfx$ of the system \eqref{eq:example1}. However, the slow dynamics $\bfp$ are hidden during training, and all training data is generated with $\bfp=p(0)=1$. Rather than explicitly update $p$, at each iterative update, the neural network is re-initialized and retrained using 10,000 datapoints simulated using the instantaneous true value of $p$ and random values of $x$ sampled from a uniform distribution from 0 to 1.

    The availability of such a volume of data across the breadth of the input space is not realistic, but is used here to illustrate the concept of iterative updates to an ANN as the underlying system evolves. The simulation results shown in \Cref{fig:boundarylayer_nn} show that the iterative scheme \eqref{eq:goveq_layereq_nn_iterative} represented in a piecewise manner by the iterative versions of the ANN model ($\Phi_{t_{0}}$, $\Phi_{t_{25}}$, $\Phi_{t_{50}}$, $\Phi_{t_{75}}$) provides an improved prediction of the values of the fast state $x$ relative to the initially trained network $\Phi_{t_0}$.

    \begin{subequations}\label{eq:example1_nn}
        \beqn
        \tilde \bfx_{k+1}& = &\phi(\hat \bfx_k, \hat \bfu_{k}, \bmp_j) \qquad \forall t \in [t_j, t_j +T_s] \\
        \tilde \bfx(t_j) & = &\tilde \bfx(t_{j-1} + T_s)
        \eeqn
    \end{subequations}

\end{example}

\subsubsection{Which parameters to update?}

When ANN models are trained, the loss function $\mathcal{L}$ is used to measure prediction error. Subsequently, the gradient of the loss function $\gradloss$ is calculated using backpropagation to approximate the contribution of each parameter to the error. Stochastic Gradient Descent-based methods then use this gradient to inform the size and direction of updates to the model parameters.

ANN models benefit from high dimensional space to escape local minima in the loss function that are distant from the global minimum during training\cite{choromanska2015loss, zhu2018anisotropic}. However, once the model has been trained, only a small subset of model parameters $\bmj \in \bmp$ may be responsible for the output (and subsequently the error) of the model prediction\cite{Frankle2018}. This is particularly true within the scope of slow changes in the parameters of the underlying physical system where structure remains unchanged. Updating all model parameters increases the search space and increases the computational complexity of the update. In addition, only updating a subset of model parameters preserves the already-trained neural network structure.

Building on the incremental update idea previously discussed, we hypothesize that only a small subset of the model parameters need to be updated in a previously trained model to maintain model accuracy. We will use $\agradloss$ to identify to parameters $\bmj$ with the largest contribution to the loss function and only update those parameters. 


In this work, we use two methods to identify the subset of parameters to be updated. Both of these methods use quantile calculations shown in \eqref{eq:quantile} where $z$ is the $q$-th quantile of the elements of $A$.

\begin{equation}
    \label{eq:quantile}
    \text{quantile}(A,q) = z
\end{equation}

\emph{Proportion-based selection} updates a fixed number of parameters at each iteration based on whether the gradient of loss for that specific parameter $\agradloss[\bmj]$ is greater than the $q$-th quantile of the elements of $\agradloss$ as shown in \eqref{eq:proportion}. This method is abbreviated as Prop.$q$.

\begin{equation}
    \label{eq:proportion}
    \text{Prop.}q = \{\bmj \in \bmp : \agradloss[\bmj] > \text{quantile}(\agradloss,q)\}
\end{equation}

\emph{Magnitude-based selection} updates a variable number of parameters at each iteration based on whether for that specific parameter $\agradloss[\bmj]$ is greater than the quantile $q$ of the gradient of the loss function calculated on the validation set during training $|\nabla_{\pi 0}\mathcal{L}_{\text{Validation}}|$ as shown in \eqref{eq:magnitude}. This method is abbreviated as Mag.$q$.

\begin{equation}
    \label{eq:magnitude}
    \text{Mag.}q = \{\bmj \in \bmp : \agradloss[\bmj] > \text{quantile}(|\nabla_{\pi 0}\mathcal{L}_{\text{Validation}}|,q)\}
\end{equation}

\subsubsection{Model Maintenance: \emph{How} to modify ANN parameters}

We propose \textbf{updating} a subset (specifically the subset $\prime$) of model parameters using the Subset Extended Kalman Filter (SEKF), in contrast to retraining or finetuning a model which follow the same basic process as training an ANN.

Kalman Filters are a class of recursive filters that were developed to estimate the state of a linear dynamic system from a series of noisy measurements\cite{kalman1960new}. The Extended Kalman Filter (EKF) is an extension of the Kalman Filter that can be used to estimate the state and update model parameters of nonlinear dynamical systems\cite{mcgee1985discovery}. The EKF uses a linear approximation of the nonlinear system dynamics and measurement functions to estimate the state of the system. The EKF was proposed as a second-order training method for ANN model parameters in 1992 by \citet{puskorius1992recurrent}. However, the computational cost increases exponentially with the number of model parameters $l$ and the number of model outputs, causing this to become an ineffective training method, compared to stochastic gradient descent-based methods, which have a linear cost per parameter.

The SEKF is a modification of the EKF that updates only the subset of model parameters $j$ that were selected in the previous step~(\ref{eq:proportion}--\ref{eq:magnitude}), rather than all parameters, to reduce the computational cost of the Kalman filter update operation, and preserve ANN model structure. The SEKF is initialized at $k=0$ as shown in \eqref{eq:sekf_init} with the initial model parameters $\pi_0$, the initial posterior covariance matrix $P_0$, the initial process noise covariance matrix $Q_0$, and the learning rate at iteration $k$ is $\eta_k$.

\begin{subequations}
    \label{eq:sekf_init}
    \beqn
    P_0 &=& p_0 I\\
    Q_0 &=& q_0 I
    \eeqn
\end{subequations}

As each new datapoint is streamed in, the SEKF performs the sequential tasks of estimation~\eqref{eq:sekf_prediction}, parameter selection~\eqref{eq:sekf_subselection}, and parameter update~\eqref{eq:sekf_update}. During estimation, the model output at time $k$ is predicted using the model parameters at time $k-1$ and the input at time $k$. The error between the predicted output and the measured output $e_k$ (refered to as the innovation in Kalman Filtering literature) is calculated. In addition, the approximation of measurement noise $R_k$ is calculated as the inverse of the learning rate $\eta_k$. Finally, the flattened Jacobian matrix $H_k$ is calculated as the gradient of the model output with respect to the model parameters at time $k-1$.

\begin{subequations}
    \label{eq:sekf_prediction}
    \beqn
    \tilde{\bfx}_{k} &=& \phi(\bfx_k, \bfu_k, \bmp_{k-1})\label{eq:sekf_prediction_a}\\
    e_k &=& \hat{\bfx}_k - \tilde{\bfx}_k\\
    R_k &=& I\frac{1}{\eta_k}\\
    H_k &=& \nabla_{\pi_k}\hat{\bfx}_{k-1}
    \eeqn
\end{subequations}

During parameter selection, the subset of parameters $\bmj$ to be updated is selected as described in (\ref{eq:proportion}-\ref{eq:magnitude}). The subset Jacobian $H_k'$, subset posterior covariance $P_{k-1}'$, subset process noise covariance $Q_k'$, and the subset measurement noise covariance $R_k'$ matrices are constructed using the indices of the selected parameters. This step is what distinguishes the SEKF from standard EKF methods, and would be identical to the EKF if $\bmj$ included all parameters $1\cdots l$.

\begin{subequations}
    \label{eq:sekf_subselection}
    \beqn
    H_k' &=& H_k [\bmj]\\
    P_{k-1}' &=& P_{k-1} [\bmj,\bmj]\\
    Q_k' &=& Q_k [\bmj,\bmj]\\
    R_k' &=& R_k [\bmj,\bmj]
    \eeqn
\end{subequations}

The update step begins with the most computationally intensive step: calculating the scaling matrix $A_k^\prime$ through matrix inversion \eqref{eq:sekf_prediction_a}. The computational cost of this step is determined by the size of the matrix inverted which is a function of the number of parameters selected $\bmj$ and the number of model outputs. The Kalman gain $K_k'$ is then calculated as the product of the subset posterior covariance matrix $P_{k-1}'$, the subset Jacobian matrix $H_k'$, and the scaling matrix $A_k'$. The subset model parameters $\bmp_k[\bmj]$ are updated by adding the product of the Kalman gain and the error $e_k$ to the value of the model parameters at time $k-1$. Finally, the subset posterior covariance matrix $P_k[\bmj,\bmj]$ is updated by subtracting the product of the Kalman gain and the subset Jacobian matrix from the subset posterior covariance matrix at time $k-1$ and adding the subset process noise covariance matrix $Q_k'$.

\begin{subequations}
    \label{eq:sekf_update}
    \beqn
    A_k' &=& [H_k'^T P_{k-1}' H_k' + R_k']^{-1}\label{eq:sekf_update_a}\\
    K_k' &=& P_{k-1}' H_k' A_k'\\
    \bmp_k[\bmj] &=& \bmp_{k-1}[\bmj] + K_k' e_k\\
    P_k[\bmj,\bmj] &=& [I - K_k' H_k'^T] P_{k-1}' + Q_k'
    \eeqn
\end{subequations}

So long as the initial model weights are trained optimally, the extended Kalman Filter should converge exponentially assuming the system is twice differentiable, and the mismatch is observable\cite{krener2003convergence}.

Kalman filtering posesses one key advantage over offline gradient-based methods: the matrix operations the Kalman filter uses have a consistent computational cost performed with each new datapoint in the datastream. In contrast, gradient descent-based methods must evaluate model prediction loss on the breadth of the training dataset in order to calculate the gradient of loss w.r.t. parameters. Gradient descent-based methods are also sensitive to hyperparameters such as learning rates and learning schedules to ensure convergence. In addition, this full pass must occur over tens to thousands or millions of epochs during the retraining or finetuning process. When maintaining an already-trained model, the computational cost of these repeated epochs can rival or far exceed the cost of the Kalman filter operations.

\section{Case Studies}

In order to evaluate updating ANN models using the SEKF, we consider the performance of retraining, finetuning, and updating on a series of ANN model applications of increasing comlexity. In all cases, the purpose of the model maintenance is to provide an accurate prediction of the future output of a dynamical system. We begin with a trained ANN model and introduce monotonically increasing drift in the underlying system and investigate how the maintenance methods compare in terms of Mean Absolute Error (MAE) as defined in \Cref{eq:mae}, Mean Squared Error (MSE) as defined in \Cref{eq:mse}, Mean Time Per Iteration (MTPI) as defined in \Cref{eq:mtpi}, and manual tuning required. We denote metrics calculated on data normalized to a mean value of 0 and variance of 1 as ``normalized'' for example: ``normalized Mean Absolute Error'' or nMAE. Forecast Skill Score (FSS) is used to benchmark the effectiveness of maintenance methods relative to no maintenance and is calculated as shown in \Cref{eq:fss}. In order to determine the effectiveness of the model maintenance methods and control for the effect of parameter selection, we apply finetuning and updating using the SEKF to the set of all model parameters as well as the subset of parameters selected using magnitude and proportion-based thresholds.

\begin{equation}
    \label{eq:mae}
    \text{MAE} = \frac{1}{n} \sum_{i=1}^{n} |\bfx_i - \tilde \bfx_i|
\end{equation}

\begin{equation}
    \label{eq:mse}
    \text{MSE} = \frac{1}{n} \sum_{i=1}^{n} (\bfx_i - \tilde \bfx_i)^2
\end{equation}

\begin{equation}
    \label{eq:fss}
    \text{FSS} = 1 - \frac{\text{Loss}_{\text{maintenance}}}{\text{Loss}_{\text{no maintenance}}}
\end{equation}

\begin{equation}
    \label{eq:mtpi}
    \text{MTPI} = (\sum_{i=1}^{n}(t_k-t_{k-1}))/n
\end{equation}

Retraining and finetuning occur when a pre-determined loss threshold is reached, and only differ in the initial values of the model parameters (either re-initialized in a Xavier normal distribution\cite{glorot2010understanding} or, using the current values). When the threshold has been reached, a retraining/finetuning event occurs, using the Adam optimizer\cite{kingma2014adam} and a fixed length rolling horizon of recent data for a fixed number of epochs or until a training-loss-threshold is satisfied.

In contrast to retraining and finetuning, which are driven by fixed events, model updating using the SEKF occurs as each new pair of true and predicted datapoints become available. As a new pair of datapoints $(\hat{\bfx}_k,\hat{\bfu}_k)$ becomes available, the SEKF updates the selected model parameters as well as the posterior matrix $P$.

To control for sensitivity to the number of parameters selected, results for each maintenance method are compared for all parameters, as well as two variants of magnitude-based parameter selection (Mag.99 and Mag.95) and two variants of proportion-based parameter selection (Prop.99 and Prop.95).

A list of the hyperparameters considered for both retraining and updating and a brief discussion on these is provided in \Cref{sec:hyperparameters} of the appendix.


\subsection{One-Dimensional System with Parameter Drift}

The system presented in \Cref{eq:example1} of \Cref{ex:simpleexample} is approximated by a neural network with one hidden layer and ten nodes of the form:

\beqn
\label{eq:ann_twotimescale}
\hat \bfx_{k+1} = \phi(\hat \bfx_k, \hat \bfu_k, \bmp)
\eeqn

The ANN predicts $\hat{\bfx}_{k+1}$ given $\hat{\bfx}_{k}$ and $\hat{\bfu}_{k}$ while $p$ evolves. Initial training occurs training the neural network up until $t=15$ with a validation set ending at $t=20$. Model maintenance methods then are applied at $t=20$ and until $t=100$. The full details of the system and training are provided in \Cref{subsec:twotimescale_details} of the appendix.


\subsection{CSTR with Reaction Rate Constant Drift}

\citet{Kumar2023a} present a CSTR model of an irreversible reaction and a subseqent reversible reaction \eqref{eq:cstr} occur. Reaction rates are defined in \eqref{eq:cstr_rate}. The model predicts product concentrations ($\hat{X}_t$) across the prediction horizon $t\in 1 \ldots n_p$ given initial concentrations ($X_0$) and the exogenous feed concentrations across the prediction horizon ($C_{Af,t}$) as an exogenous input ($U$). \Cref{fig:CSTR_prediction} shows an example of the ANN inputs and outputs.

\begin{subequations}
    \label{eq:cstr}
    \beqn
    A& \overset{r_1}{\rightarrow} &B \\
    3B& \overset{r_2}{\leftrightarrow} &C
    \eeqn
\end{subequations}

\begin{subequations}
    \label{eq:cstr_rate}
    \beqn
    r_1& = &k_1 C_A \\
    r_2& = &k_{2f} C_B^3 - k_{2r} C_C  \label{eq:cstr_rate2}
    \eeqn
\end{subequations}

We use a Neural ODE (NODE)\cite{chen2018neural} architecture similar to that of \citet{Kumar2023a} shown in \Cref{fig:NODE}. Standard ANN architecture elements (shown in orange) produce a rate of change for each state that is integrated using a Runge-Kutta method recursively as the initial conditions for the next element in the prediction until the prediction horizon has been reached.

\begin{figure*}[t!]
    \centering
    \begin{subfigure}[t]{0.57\textwidth}
        \centering
        \includegraphics[width=\textwidth]{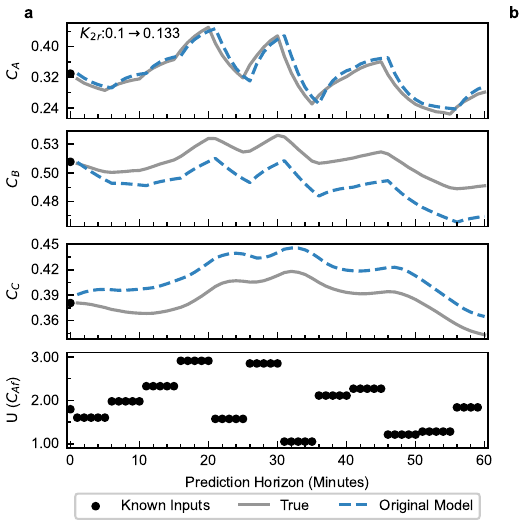}
    \end{subfigure}
    \begin{subfigure}[t]{0.42\textwidth}
        \centering
        \includegraphics[width=\textwidth]{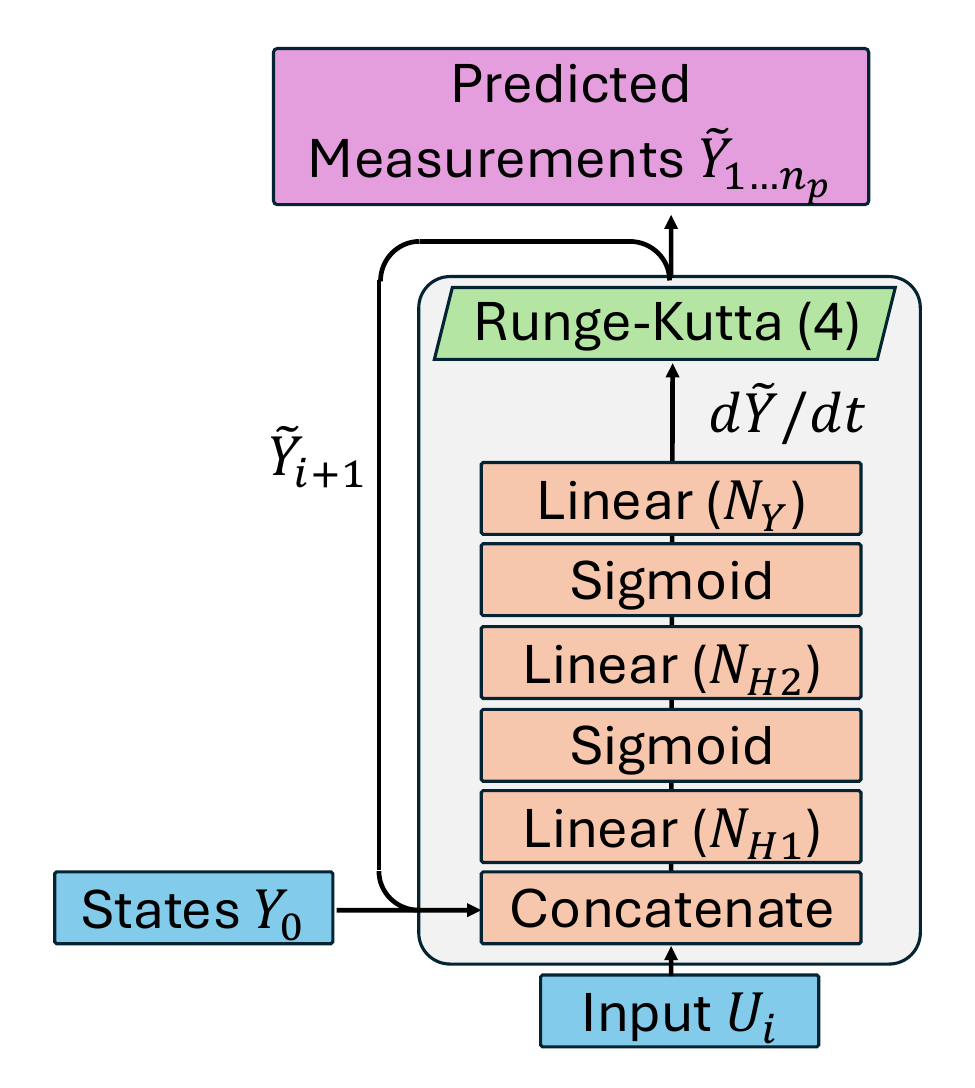}
    \end{subfigure}
    \caption{\textbf{a} The system dynamics shown in grey drift away from the original dynamics as shown by the disparity between the true concentrations and the prediction of the original model in dashed blue. \textbf{b} NODE architecture used.}
    \phantomsubfigure{fig:CSTR_prediction}
    \phantomsubfigure{fig:NODE}
\end{figure*}

Once the NODE model is trained, we introduce drift by increasing $k_{2r}$ in \Cref{eq:cstr_rate2} linearly by 0.03\% of the original value per minute and evaluate model maintenance methods over 36 hours. The full details of the model maintenance setup are provided in \Cref{subsec:cstr_details} of the Appendix. \Cref{fig:CSTR_prediction} shows an example of the predictions made by the originally trained model as $k_{2r}$ has drifted from the initial value of 0.1 to 0.13.

\subsection{Simulated Type-II Diabetic Patient with Insulin Sensitivity Drift}

\citet{palma2013estimation} describes a model of glucose and insulin dynamics in a type II diabetic patient. Ingested glucose ($D$) passes through two intermediate digestion stages--$q_1$ and $q_2$--before being absorbed in the gut as $G_{gut}$. Delivered insulin ($U$) increases the ammount of insulin in the blood plasma ($I$). Blood glucose ($G$) changes as a function of the difference between the basal glucose level ($G_b$), remote insulin absorbed by tissue ($X$), and $G_{gut}$. An example day of data is shown in \Cref{fig:glucose_exampleday}.


\begin{subequations}
    \label{eq:glucose}
    \beqn
    \frac{dG(t)}{dt}& = &-p_1(G(t)-G_b)-s_iX(t)G(t)+\frac{fk_{abs}}{V_G}G_{gut} \label{eq:glucose_1}\\
    \frac{dX(t)}{dt}& = &-p_2(X(t) - I(t) - I_b) \\
    \frac{dI(t)}{dt}& = &U(t) - k_eI(t) \\
    \frac{dq_1(t)}{dt}& = &D(t)-k_{emp}q_1(t) \\
    \frac{dq_2(t)}{dt}& = &k_{emp}(q_1(t)-q_2(t)) \\
    \frac{dG_{gut}(t)}{dt}& = &k_{emp}q_2(t)-k_{abs}G_{gut}(t)
    \eeqn
\end{subequations}

Healthy blood glucose levels range from 70--125 mg/dL (4-10 mmol/L) with more-severe consequences for hypoglycemic levels (below 70 mg/dL) \cite{bilous2021handbook}. A simple control heuristic approximates a patient's management of blood glucose: if blood sugar is below 75 mg/dL, no insulin is administered for the next hour. If blood sugar is below 60 mg/dL, a glucose supplement is ingested at a rate of 2 g/minute. \Cref{fig:glucose_distribution} shows the effect this control heuristic has on the distribution of blood glucose levels.

Five years of data are used to train a neural network to predict the states $\bfx\in\left\{G,X,I,q_1,q_2,G_{gut}\right\}$ over a prediction horizon of one hour given the inputs $\bfu\in\left\{U(t), D(t)\right\}$, and the current conditions of the states at $t=0$. After training, we evaluate model maintenance as the patient insulin sensitivity $s_i$ in \Cref{eq:glucose_1} decreases from 0.5 to 0.3 over a year. We note that this change is likely much higher than what would be encountered in practice and is introduced for the purpose of evaluating the ability of the proposed model maintenance technique.

\subsection{Fluid Catalytic Cracker Unit with Process Changes}

\citet{Santander2022b} developped a model for a Fluid Catalytic Cracker (FCC) unit---which performs the core operations in a petrolium refinery---that includes ten unit operations including feed preheating, catalytic reaction and regeneration, and seperation of product streams. \Cref{fig:FCC_variables} shows the process flow diagram of the FCC process and the measured variables. \citet{Kumar2022} used this model to create seven datasets of process operations data: Two datasets consist of normal operating conditions: one with two days of constant feed flow rate, and another with seven days of varying feed flowrates. The remaining five datasets each contain one day of operating data as a change in the underlying process is simulated. A summary of the datasets is provided in \Cref{tab:fcc_datasets}.

We train a NODE with 2 hidden layers with 64 neurons each to predict the 31 dependent variables $X$ (shown in grey in \Cref{fig:FCC_variables}) across the prediction horizon $t\in 1 \ldots 60$ given the initial conditions of the states $X_0$ and the 15 exogenous inputs $U$ (shown in green in \Cref{fig:FCC_variables}) using the normal operating conditions datasets. Next we observe the performance of the model updating methods over the course of each of the five faulty operations datasets. The full details of the FCC model and training are provided in \Cref{subsec:FCC_details} of the appendix.

\section{Results}\label{sec:results}

\subsection{One-Dimensional System with Parameter Drift}\label{subsec:twotimescale_results}

The lowest error result for each maintenance method is shown in \Cref{fig:twotimescale_lineplot}. Overall results from maintaining the system are shown in \Cref{fig:twotimescale_scatter}  with a more-complete table of results included in \Cref{tab:twotimescale_results}.

\begin{figure}
    \centerline{\includegraphics[width=183mm]{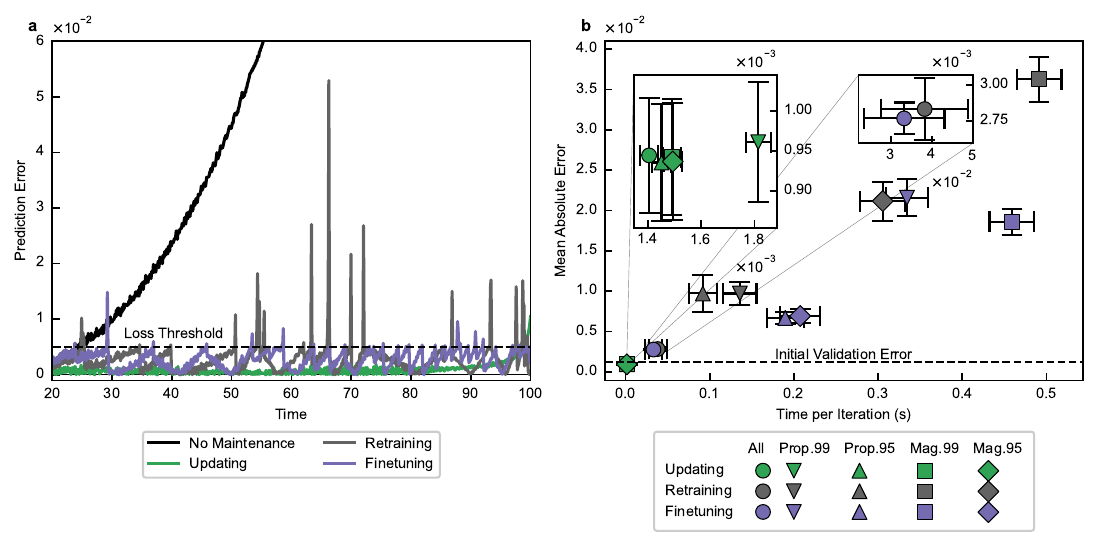}}
    \phantomsubfigure{fig:twotimescale_lineplot}
    \phantomsubfigure{fig:twotimescale_scatter}
    \caption{Results for the One-Dimensional System with Parameter Drift. \textbf{a} the lowest-error result of each maintenance method for the test data. \textbf{b} summarized results for the One-Dimensional System with Parameter Drift with 95\% confidence bounds for each maintenance method operating on all parameters, and the best result on a subset of parameters.}
\end{figure}

Both figures show dramatic differences in the error, time required per iteration, and the variance in time required per iteration between the online updates provided by the SEKF, and the offline methods (retraining and finetuning). Online updates on the incomming datastream provide consistently lower error values in contrast to the offline methods(\Cref{fig:twotimescale_lineplot}). Offline methods required significantly longer to perform maintenance than the SEKF and generally had higher error than updating. Retraining and finetuning all parameters achieved similar results and were the best results for offline methods, however all five updating methods outperformed the offline methods.

Through the three remaining case studies, we only compare updating and finetuning, since the latter generally exhibited the better computational performance of the two offline methods considered. 

\subsection{CSTR with Reaction Rate Constant Drift}\label{subsec:cstr}

\Cref{fig:CSTR_scatter} and \Cref{tab:CSTR_results} show the MSE of predictions made by updated and finetuned models as drift increases across the 60 minute time horizon shown in \Cref{fig:CSTR_prediction}. As before, model updating achieves lower error and results with less variance in prediction errors than retraining. Interestingly, when \emph{finetuning}, the time required for model maintenance increases as fewer parameters are finetuned; Conversely, when updating model parameters using the SEKF, the time required for model maintenance decreases as fewer parameters are updated and in the case of proportion-based parameter selection, without an increase in prediction error. Overall, updating achieves a lower nMSE and MTPI than finetuning.

\begin{figure}[ht]
    \centering
    \centerline{\includegraphics[width=183mm]{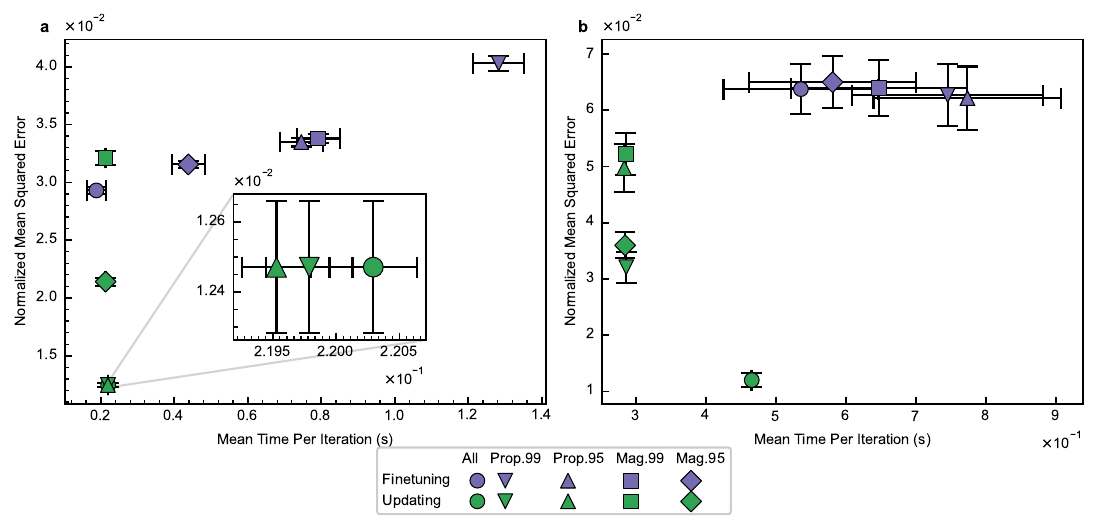}}
    \phantomsubfigure{fig:CSTR_scatter}
    \phantomsubfigure{fig:glucose_scatter}
    \caption{\textbf{a} summary statistics for the CSTR with Reaction Rate Constant Drift \textbf{b} summary statistics for the Simulated Type-II Diabetic Patient with Insulin Sensitivity Drift. In both cases, the SEKF outperforms finetuning and is able to update a subset of model parameters without significant loss in performance (if any) and in the case of the larger model of the diabetic patient, operating on a subset of model parameters causes a significant decrease in the time required to perform model maintenance.}
\end{figure}

\subsection{Simulated Type-II Diabetic Patient with Insulin Sensitivity Drift}\label{subsec:glucose}

Results shown in \Cref{fig:glucose_scatter}, and \Cref{tab:glucose_results}. As with previous systems, updating with the SEKF achieved lower error and required less time per new datapoint to maintain the model compared to retraining. As the neural network dimension increases, updating all parameters using the SEKF begins to have an effect on MTPI. However, updating all parameters did achieve the lowest nMSE of all maintenance and parameter selection methods.

\subsection{Fluid Catalytic Cracker Unit with Process Changes}\label{subsec:fcc}

\begin{figure}[ht]
    \centering
    \centerline{\includegraphics[width=183mm]{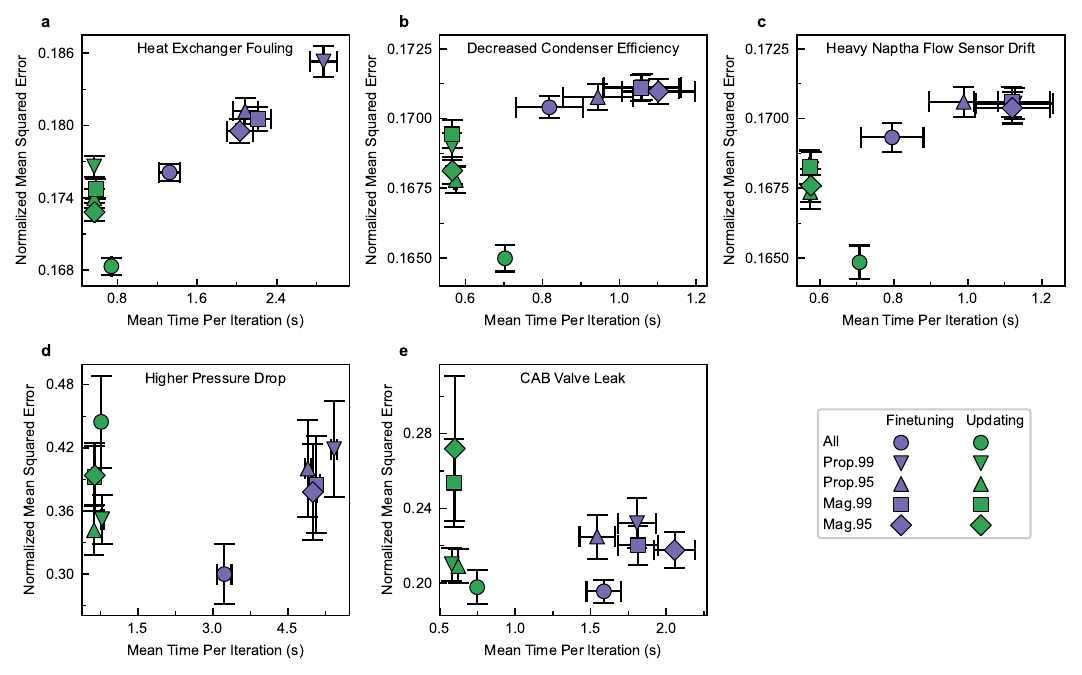}}
    \caption{\label{fig:FCC_scatter} \textbf{a}-\textbf{e} mean nMSE and MTPI as well as 95\% confidence intervals for process change Fluid Catalytic Cracker Unit system.}
\end{figure}

The results from the FCC model maintenance are shown in \Cref{fig:FCC_scatter} with each plot displaying the results of a different process change. In all cases, updating using the SEKF required less time--always staying below the one-second per update iteration level. Since updating consists of the same matrix multiplications with each new datapoint, the MTPI remained constant between scenarios. In contrast, the higher pressure drop and CAB valve leak scenarios changed significantly from the initial conditions (the unmaintained model error was 1-2 orders of magnitude higher for these cases than the others in \Cref{tab:fcc_results_1} and \Cref{tab:fcc_results_2}). In these scenarios, finetuning required significantly more time than updating and than finetuning in other scenarios. However, with the frequency of these finetuning events, and the nature of the drift (occurring drastically but then remaining constant at a new value), finetuning was able to achieve similar nMSE to updating even outperforming updating in the higher pressure drop scenario.

\section{Discussion}\label{sec:discussion}

The problem of absorbing real-time data into data-driven models is an open research question \cite{verwimp2023continual}. Considering again machine learning models as a means to map system inputs to outputs, model adaptation must balance learning new relationships \cite{dohare2024loss} with  preserving knowledge of the input/output system behavior that is still relevant\cite{kirkpatrick2017overcoming}. Ideally, the model adaptation and maintenance process would require as little human intervention as possible. The proposed approach, based on the Kalman Filter, enables online learning by \textit{updating} model parameters and covariance estimates without the need for tuning hyperparameters (such as learning rates, and choice of data to use during retraining) involved in gradient descent-based \textit{retraining} methods.

The benefits of the \emph{subset} Kalman Filter approach proposed here are twofold. First, it reduces computational cost by reducing the size of the matrix involved in the inversion operation (equation \ref{eq:sekf_update_a}) - the computational cost of the inversion operation scales with the cube of the dimension of the matrix. The computational advantage of the proposed approach became apparent as the dimensions (and number of parameters) of the neural networks increased with increased system complexity (see \Cref{tab:models}) in the four case studies considered. The results in Figures \ref{fig:twotimescale_scatter}, \ref{fig:CSTR_scatter}, \ref{fig:glucose_scatter} and \ref{fig:FCC_scatter} show an increasing discrepancy between the time required to update all the model parameters using the KF, and updating a parameter subset using the SEKF. Additionally, the discrepancy between the time (and computational effort) needed by (SE)KF and the time required for model finetuning grows as the model size increases, as seen in the aforementioned figures.

Second, updating a subset of the model parameters is expected to yield a better balance between the plasticity needed to adapt to new circumstances and the stability needed to maintain overall model performance. Under the circumstances considered (dealing with a slow, typically monotonic parameter drift in the underlying physical system), the overall functional structure or overall input/output behavior predictions of the model are unlikely to change. Updating lends itself naturally to periodic updates of a small number of parameters in the model.

\begin{table}
    \caption{Characteristics of the models used in the case studies}\label{tab:models}
    \begin{tabular}{llrrr}
        \toprule
        Case Study               & Model Description     & \# Inputs & \# Outputs & \# Parameters \\
        \midrule
        One-Dimensional System with Parameter Drift       & 1-layer NN (10)       & 2         & 1          & 41            \\
        Reactor with Rate Constant Drift & 2-Layer NODE (16, 16) & 63        & 180        & 419           \\
        Simulated Diabetic Patient         & 2-Layer NODE (32, 32) & 126       & 360        & 1,574         \\
        FCC-Fractionator & 2-Layer NODE (64, 64) & 931       & 1,860      & 9,247         \\
        \bottomrule
    \end{tabular}
\end{table}

Utilizing the gradient of loss as the parameter selection criterion for model updating provides an opportunity to visualize the effects of system parameter drift on model accuracy. We found these visualizations rather insightful and illustrate them for the case of the one-dimensional system with parameter drift. \Cref{fig:two_timescale_gradient} shows the change in the gradient of the loss function $\gradloss$ as a function of parameter $p$. Our hypothesis that it is sufficient to update only a subset of the model parameters is supported by  the observation that the loss is not uniformly impacted by all parameters. Rather, the gradient of the loss function is high for only a few parameters, and is practically indifferent to many others.  This suggests that the mismatch between the model predictions and the outputs of the underlying system is can be traced to a small subset of the model parameters (and can be remedied by updating these).

\begin{figure}
    \centerline{\includegraphics[width=183mm]{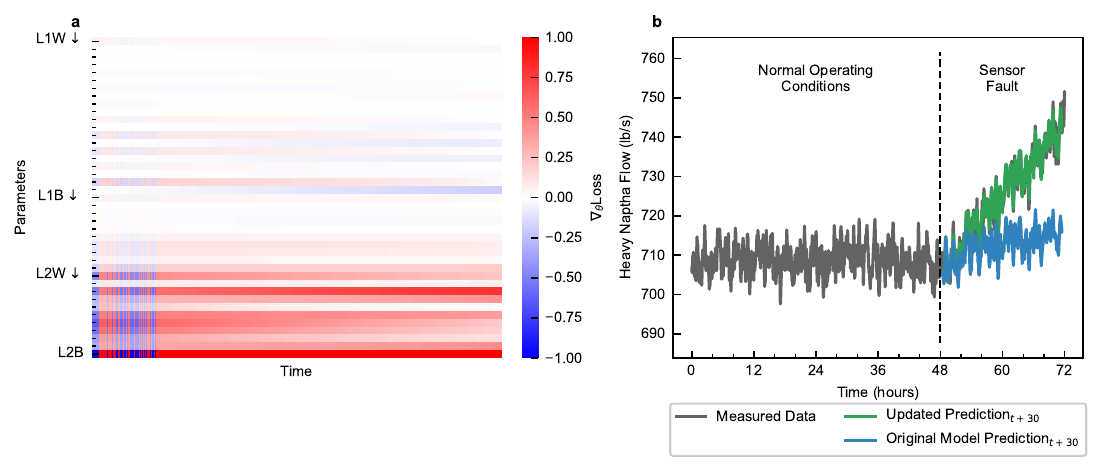}}
    \caption{Two  observations when maintaining NN models. \textbf{a} the gradient of loss is high for a subset of model parameters as drift increases. Labels L1 and L2 refer to layer 1 and layer 2 of the neural network and W and B refer to the weight and bias paramters (One-Dimensional System with Parameter Drift). \textbf{b} the updating process accommodates a faulty sensor measurement in the FCC-Fractionator system. Model updating should also account for fault detection and data validation so that models track the true underlying system}
    \phantomsubfigure{fig:two_timescale_gradient}
    \phantomsubfigure{fig:HN_drift}
\end{figure}


\subsection{Comparison with Literature}



The vast majority of the work available in updating machine learning models of physical systems uses finetuning with gradient descent methods. The authors rely on updated data sets that are collected on a relatively short horizon that shifts in time (``rolling horizon''). \citet{bhadriraju2019machine} discuss updates to some of the parameters of dynamic ML models derived via symbolic regression when the prediction error crosses a pre-specified threshold. The parameter selection is driven by regularization terms included in the objective function.


When updating ANN models of physical systems, authors tend to use finetuning calculations similar to those described in this work \cite{zheng2023physics, zheng2022online, hu2023online}, if updating is considered at all. Additionally, we are not aware of any prior case study that probes updating models of practically relevant dimensions and complexity. The aforementioned reports comprise a single case study, with all -- with the exception of \cite{drgovna2021physics} -- considering a simple CSTR or batch reactor.

The use of the EKF to train ANN models has been proposed early on by, e.g., \citet{puskorius2001parameter}. The use of EKF for ANN model updating (particularly in a classification context) has been reported more recently. The authors identified computational cost as a key challenge, and proposed, e.g., matrix decomposition \cite{chang2022diagonal}, Monte Carlo methods \cite{ghosh2016assumed, wagner2023kalman} buffer methods \cite{nguyen2017variational,kurle2019continual}, and representing the posterior precision using a diagonal plus low-rank approximation \cite{chang2023low,mishkin2018slang, lambert2022recursive, agarwal2018case, min2022one, ritter2018online,daxberger2021laplace} to reduce computational cost.

In contrast, our proposed approach is predicated on parameter updating in the original ANN model parameter space. Its novelty  consists of  recognizing that only a subset of the parameters of a \textit{trained} model contribute to the predictions (and the corresponding error) at a given time point. Thus, the SEKF updates only a subset of the parameters. Finding these parameters based on the gradient of loss is another novel contribution of this work. The SEKF lowers the computational cost of the update operation. We additionally posit that updating only a subset of model parameters affords some permanence to the information embedded in the model and provides the ``long-term memory'' effect discussed e.g., by \citet{dohare2024loss}.


Incremental or continual learning is a related field in machine learning of contemporary interest. \citet{van2022three} provide a concise taxonomy for incremental learning problems based how the task the model performs changes over time and whether new tasks are introduced to the same model. Two key distinctions between this work and the bulk of the existing continual learning literature are:

\begin{enumerate}
    \item Much of the continual learning literature is concerned with adapting the same model to new tasks while maintaining performance on past examples\cite{van2024continual,galashov2024non, zeng2019continual,imam2020rapid}. In contrast, we are only concerned with current and future model performance, not about preserving performance on past data. Thus, in this work, we are not concerned with the performance of the model at time $k$ on data from time $0 \ldots k-1$. Instead, we are only concerned  about the performance of the model at time $k$ on data from time $k$.
    \item The overwhelming majority of continual learning literature addresses image-classification problems, with the only examples to the contrary being \citet{chang2023low} who learn time-series regression online, but do not explicitly address drift in the system, and the toy problem of bit-flipping shown by \citet{dohare2024loss}. In contrast, we introduce  regression problems that explicitly model drift in the underlying dynamical system.. In our experience, this better reflects the real-world environments that models must be maintained in.
\end{enumerate}

It is worth noting that another partially related field--transfer learning--often takes advantage of subselecting model parameters to adapt existing models to new environments or data. For example, for many deep learning classification models, only the final layer is retrained or finetuned \cite{chollet2021deep}. However, this approach is heuristic and may not address the changes in how the underlying physical system evolves. 

\section{Conclusion}\label{sec:conclusion}

Automated maintenance of ANN models in the presence of drift in the behavior of the underlying physical dynamical system is a challenging problem that has not been widely studied in the literature. Inspired by multi-scale modeling approaches, and justified theoretically by singular pertubation analysis, we propose using the SEKF to update ANN model parameters identified by the $\agradloss$ in real-time. The online nature of the Kalman filtering algorithm exploits the known neural network structure to identify the parameters to update and circumvents the subjective and challenging task of distinguishing between relevant and  irrelevant training data. In the age of large language models and models with trillions of parameters, stateful model updating may allow machine learning models to be applied in a more data-efficient and energy efficient manner without sacrificing performance.


Future work should challenge the current assumptions of monotonic drift and unchanged system structure. This may include structural changes to the underlying system, transfer learning to adapt to similar but different systems, or incorporating re-initialization of certain model parameters when persistent poor performance is observed to re-identify the model structure.

Finally, practical applications should consider the implementation of model updating as part of a control loop. This will require consideration of system stability. Additionally, \Cref{fig:HN_drift} shows a potentially pathological situation where a sensor of the FCC system drifts. Thus, an increased discrepancy between model predictions and new data may not be due to a change in the underlying system, but instead due to a fault in the data collection process. Model updating accommodates the faulty sensor measurement by predicting the faulty readings. In practice, model updating should occur in conjunction with fault detection and data validation so that model updates track the true underlying system. We hypothesize that information contained in the $\gradloss$ could assist with fault detection and isolation, and this is a matter we will consider in future research.


\backmatter

\section*{Acknowledgements}
The authors would like to acknowledge the helpful feedback of industry sponsors in the Texas Wisconsin California Control Consortium (TWCCC)
\subsection*{Author contribution}
\textbf{Joshua Hammond}: Methodology, Software, Formal analysis, Investigation, Writing - Original Draft, Writing - Review \& Editing, Visualization
\textbf{Tyler Soderstrom}: Resources, Writing - Review \& Editing, Supervision, Funding acquisition
\textbf{Brian Korgel}: Resources, Writing - Review \& Editing, Supervision
\textbf{Michael Baldea}: Conceptualization, Methodology, Resources, Writing - Original Draft, Writing - Review \& Editing, Supervision, Project administration, Funding acquisition
\subsection*{Funding}
Financial support from ExxonMobil is acknowledged with gratitude.
\subsection*{Conflict of interest/Competing interests}
The authors declare no conflicts of interest
\subsection*{Code availability}
The code used to generate data, train models, update/retrain/finetune models, create figures, as well as the generalized SEKF optimizer is available at \url{https://github.com/joshuaeh/Subset-Extended-Kalman-Filter/tree/v2025.02.25} under an MIT license.
\subsection*{Data availability}
The data are available at \url{https://doi.org/10.18738/T8/VA0SQI} \cite{Hammond2025a}




\begin{appendices}

    \section{Hyperparameters}\label{sec:hyperparameters}

    Hyperparameters tat must be tuned during training, retraining and finetuning include the following. For the sake of brevity, we will refer to all of these processes as ``training'':
    \begin{itemize}
        \item Optimizer learning rate \& associated parameters such as momentum and learning rate schedule if applicable
        \item Loss threshold for training
        \item Training epochs, batch size, and early stopping criteria
        \item Length of rolling horizon of data to-be-used in training
        \item Parameter selection method \& threshold
    \end{itemize}

    Similarly, when updating model parameters using the SEKF, the following hyperparameters must be tuned:

    \begin{itemize}
        \item Kalman Filter ``learning rate'' $\eta$
        \item Initial posterior covariance matrix $p_0$ and process noise covariance matrix $q_0$
        \item Parameter selection method \& threshold
    \end{itemize}

    Following \cite{puskorius2001parameter}, we used the values: $p_0:=100$, $q_0:=0.1$ and $\eta_0:=0.001$.

    When comparing the two methods, one of the biggest advantages of online methods that operate on the incomming datastream is the elimination of the hyperparameters to do with when training should occur, how long it should occur, and distinguishing between data that should and should not be included in that training process.

    \section{Case Study Details}\label{sec:details}

    \subsection{Setup}\label{subsec:setup}

    Reported results are from a machine with an AMD Ryzen 9 5950X 16-Core Processor, 64 GB of RAM, and an NVIDIA GeForce RTX 3090 Ti GPU. All computations in this work were performed using Python 3.10.4 and PyTorch 2.4.0.

    \subsection{One-Dimensional Parameter Drift System}\label{subsec:twotimescale_details}

    \textbf{Model Architecture:} Two inputs, one hidden layer with 10 neurons and Sigmoid nonlinearity, and one output.

    \textbf{Initial Training:} 150 datapoints for training, 50 for validation. The Adam optimizer is used with an initial learning rate of 0.01, Xavier initialization, and Mean Absolute Error loss objective function for 1,000 epochs and minibatches of 5 samples. Learning rate is reduced using the \texttt{ReduceLROnPlateau} scheduler with a scaling factor of 0.3 and patience of 50 epochs.

    \textbf{Model Maintenance:} As the system evolves in a convcave shape from $t=20$ to $t=100$, we observe the Mean Absolute Error (due to the sensitivity with small numbers) using retraining, finetuning, and updating using the SEKF and online adam. For each maintenance method, we compare the MAE and time taken to perform maintenance while varying the parameter selection method and learning rate. A table of all the hyperparameters used is shown in \Cref{tab:twotimescale_hyperparameters}, and the results are shown in \Cref{tab:twotimescale_results}.

    \begin{table}
        \caption{Hyperparameters for One-Dimensional Parameter Drift System}\label{tab:twotimescale_hyperparameters}
        \begin{tabular}{ll}
            \toprule
            Hyperparameter             & Values                                      \\
            \midrule
            Retraining Loss Threshold  & 0.1                                         \\
            Retraining Epochs          & 50                                          \\
            Learning Rate              & 0.1, 0.01, 0.001, 0.0001                    \\
            SEKF $p_0$                 & 100                                         \\
            SEKF $q_0$                 & 0.1                                         \\
            Parameter Selection Method & All Parameters                              \\
            \hspace{2cm}               & Prop.99, Prop.95, Prop.90, Prop.75, Prop.50 \\
            \hspace{2cm}               & Mag.99, Mag.95, Mag.90, Mag.75, Mag.50      \\
            \bottomrule
        \end{tabular}
    \end{table}
    \begin{table}
        \caption{Results for One-Dimensional Parameter Drift System}\label{tab:twotimescale_results}
        \begin{tabular}{lllrrrr}
            \toprule
            Method         & Parameters     & Best LR & MAE                    & 95\% CI                & MTPI (s)               & 95\% CI                \\
            \midrule
            Training       & -              & -       & $7.1722\times10^{-04}$ & $1.5799\times10^{-04}$ & -                      & -                      \\
            Validation     & -              & -       & $1.2223\times10^{-03}$ & $1.8203\times10^{-04}$ & -                      & -                      \\
            \midrule
            No Maintenance & -              & -       & $1.2157\times10^{-01}$ & $8.5764\times10^{-03}$ & -                      & -                      \\
            \midrule
            SEKF           & All Parameters & 0.1     & $9.4449\times10^{-04}$ & $7.2069\times10^{-05}$ & $1.4035\times10^{-03}$ & $3.3519\times10^{-05}$ \\
            SEKF           & Prop.99        & 0.1     & $9.6102\times10^{-04}$ & $7.5349\times10^{-05}$ & $1.8146\times10^{-03}$ & $4.6765\times10^{-05}$ \\
            SEKF           & Prop.95        & 0.1     & $9.3555\times10^{-04}$ & $7.3076\times10^{-05}$ & $1.4503\times10^{-03}$ & $3.4497\times10^{-05}$ \\
            SEKF           & Prop.90        & 0.1     & $9.3655\times10^{-04}$ & $7.1791\times10^{-05}$ & $1.5893\times10^{-03}$ & $3.8147\times10^{-05}$ \\
            SEKF           & Prop.75        & 0.1     & $9.4289\times10^{-04}$ & $7.1893\times10^{-05}$ & $1.5173\times10^{-03}$ & $3.5665\times10^{-05}$ \\
            SEKF           & Prop.50        & 0.1     & $9.4427\times10^{-04}$ & $7.2033\times10^{-05}$ & $1.4599\times10^{-03}$ & $3.3312\times10^{-05}$ \\
            SEKF           & Mag.99         & 0.1     & $9.4221\times10^{-04}$ & $7.2946\times10^{-05}$ & $1.4900\times10^{-03}$ & $3.2468\times10^{-05}$ \\
            SEKF           & Mag.95         & 0.1     & $9.3681\times10^{-04}$ & $7.3101\times10^{-05}$ & $1.4929\times10^{-03}$ & $3.4344\times10^{-05}$ \\
            SEKF           & Mag.90         & 0.1     & $9.4202\times10^{-04}$ & $7.2870\times10^{-05}$ & $1.7419\times10^{-03}$ & $4.4089\times10^{-05}$ \\
            SEKF           & Mag.75         & 0.1     & $9.4301\times10^{-04}$ & $7.1901\times10^{-05}$ & $1.5054\times10^{-03}$ & $3.4589\times10^{-05}$ \\
            SEKF           & Mag.50         & 0.1     & $9.4425\times10^{-04}$ & $7.2030\times10^{-05}$ & $1.5291\times10^{-03}$ & $3.4857\times10^{-05}$ \\
            \midrule
            Retraining     & All Parameters & 0.01    & $2.8311\times10^{-03}$ & $2.1368\times10^{-04}$ & $3.8187\times10^{-02}$ & $1.0582\times10^{-02}$ \\
            Retraining     & Prop.99        & 0.1     & $9.7086\times10^{-03}$ & $1.4206\times10^{-03}$ & $1.3580\times10^{-01}$ & $1.9778\times10^{-02}$ \\
            Retraining     & Prop.95        & 0.1     & $9.7200\times10^{-03}$ & $2.3068\times10^{-03}$ & $9.1777\times10^{-02}$ & $1.6843\times10^{-02}$ \\
            Retraining     & Prop.90        & 0.001   & $1.5510\times10^{-02}$ & $2.5220\times10^{-03}$ & $1.5244\times10^{-01}$ & $2.0523\times10^{-02}$ \\
            Retraining     & Prop.75        & 0.01    & $1.0688\times10^{-02}$ & $1.4904\times10^{-03}$ & $1.6533\times10^{-01}$ & $2.0776\times10^{-02}$ \\
            Retraining     & Prop.50        & 0.01    & $4.5568\times10^{-03}$ & $6.4040\times10^{-04}$ & $8.7818\times10^{-02}$ & $1.6210\times10^{-02}$ \\
            Retraining     & Mag.99         & 0.01    & $3.6247\times10^{-02}$ & $2.8009\times10^{-03}$ & $4.9133\times10^{-01}$ & $2.6605\times10^{-02}$ \\
            Retraining     & Mag.95         & 0.1     & $2.1158\times10^{-02}$ & $2.4157\times10^{-03}$ & $3.0559\times10^{-01}$ & $2.6554\times10^{-02}$ \\
            Retraining     & Mag.90         & 0.01    & $1.1502\times10^{-02}$ & $1.6735\times10^{-03}$ & $1.3825\times10^{-01}$ & $2.0309\times10^{-02}$ \\
            Retraining     & Mag.75         & 0.01    & $9.0816\times10^{-03}$ & $1.3442\times10^{-03}$ & $1.4307\times10^{-01}$ & $2.0416\times10^{-02}$ \\
            Retraining     & Mag.50         & 0.01    & $3.6728\times10^{-03}$ & $6.2640\times10^{-04}$ & $6.2969\times10^{-02}$ & $1.4322\times10^{-02}$ \\
            \midrule
            Finetuning     & All Parameters & 0.01    & $2.7663\times10^{-03}$ & $1.1058\times10^{-04}$ & $3.3159\times10^{-02}$ & $9.8740\times10^{-03}$ \\
            Finetuning     & Prop.99        & 0.1     & $2.1578\times10^{-02}$ & $2.3114\times10^{-03}$ & $3.3439\times10^{-01}$ & $2.5342\times10^{-02}$ \\
            Finetuning     & Prop.95        & 0.01    & $6.6838\times10^{-03}$ & $7.7625\times10^{-04}$ & $1.8971\times10^{-01}$ & $2.2005\times10^{-02}$ \\
            Finetuning     & Prop.90        & 0.01    & $3.7946\times10^{-03}$ & $3.3264\times10^{-04}$ & $8.3607\times10^{-02}$ & $1.5796\times10^{-02}$ \\
            Finetuning     & Prop.75        & 0.01    & $3.3549\times10^{-03}$ & $2.9215\times10^{-04}$ & $7.4349\times10^{-02}$ & $1.5052\times10^{-02}$ \\
            Finetuning     & Prop.50        & 0.01    & $3.1643\times10^{-03}$ & $1.4219\times10^{-04}$ & $7.8673\times10^{-02}$ & $1.5422\times10^{-02}$ \\
            Finetuning     & Mag.99         & 0.1     & $1.8599\times10^{-02}$ & $1.6143\times10^{-03}$ & $4.5908\times10^{-01}$ & $2.6662\times10^{-02}$ \\
            Finetuning     & Mag.95         & 0.01    & $6.9220\times10^{-03}$ & $8.8790\times10^{-04}$ & $2.0746\times10^{-01}$ & $2.3531\times10^{-02}$ \\
            Finetuning     & Mag.90         & 0.01    & $6.8710\times10^{-03}$ & $8.1431\times10^{-04}$ & $1.7478\times10^{-01}$ & $2.2201\times10^{-02}$ \\
            Finetuning     & Mag.75         & 0.01    & $3.9967\times10^{-03}$ & $3.4413\times10^{-04}$ & $9.7710\times10^{-02}$ & $1.7439\times10^{-02}$ \\
            Finetuning     & Mag.50         & 0.01    & $2.8893\times10^{-03}$ & $1.4519\times10^{-04}$ & $6.2236\times10^{-02}$ & $1.4270\times10^{-02}$ \\
            \bottomrule
        \end{tabular}
    \end{table}

    \subsection{CSTR with Reaction Rate Drift}\label{subsec:cstr_details}

    \textbf{Model Architecture:} NODE with 3 states, 1 exogenous input, 2 hidden layers with Sigmoid nonlinearity and 16 neurons each.

    \textbf{Training procedure:} Parameters shown in \Cref{tab:CSTR_parameters} and training procedure were borrowed from \citet{Kumar2023a}. 36 hours of data were simulated with 24 hours being used for training, 6 for validation, and 6 for testing. Training occured with the Adam optimizer and batch size of 100 samples and 150 epochs. Data was scaled using the StandardScaler from scikit-learn\cite{scikit-learn} to set the mean to 0 and variance to 1.

    \begin{table}
        \caption{Parameters for Reaction Parameter Drift Study}
        \label{tab:CSTR_parameters}
        \begin{tabular}{lrr}
            \toprule
            Parameter            & Value            & Units                   \\
            \midrule
            $F$                  & 0.6              & m$^3$/min               \\
            $V$                  & 15.0             & m$^3$                   \\
            $C_{A,f}$            & [1.0, 3.0]       & mol/m$^3$               \\
            $k_1$                & $2\times10^{-1}$ & mol/(m$^3\cdot$min)     \\
            $k_{2f}$             & $5\times10^{-1}$ & mol$^6$/(m$^2\cdot$min) \\
            $k_{2r, 0}$          & $1\times10^{-1}$ & mol/(m$^3\cdot$min)     \\
            $\frac{dk_{2r}}{dt}$ & $3\times10^{-5}$ & mol/(m$^3\cdot$min)     \\
            \bottomrule
        \end{tabular}
    \end{table}

    \textbf{Model Maintenance:} While $k_{2r}$ drifts over the 36-hour maintenance window, we maintain the model using finetuning and updating using the SEKF. The SEKF was initialized with learning rate 3e-4, q=0.1, and $P_0=100\cdot I$. Finetuning occured using the Adam optimizer and a learning rate of 1e-3 and 50 retraining epochs. The finetuning dataset was comprised of up to the last 50 minutes of data. The results are shown in \Cref{tab:CSTR_results}.

    \begin{table}
        \caption{Reaction parameter drift results}
        \label{tab:CSTR_results}
        \begin{tabular}{llrrrrrr}
            \toprule
            Method         & Params.   & MSE                    & 95\%CI                 & FSS                    & 95\%CI                 & MTPI (s)               & 95\%CI                 \\
            \midrule
            Training       & -         & 8.$8168\times10^{-02}$ & 5.$1032\times10^{-02}$ & -                      & -                      & -                      & -                      \\
            Validation     & -         & 9.$9257\times10^{-02}$ & 6.$4142\times10^{-02}$ & -                      & -                      & -                      & -                      \\
            \midrule
            No Maintenance & -         & 7.$3629\times10^{-01}$ & 2.$5886\times10^{-02}$ & -                      & -                      & -                      & -                      \\
            \midrule
            Finetuning     & All       & 2.$9299\times10^{-02}$ & 2.$9497\times10^{-04}$ & 8.$5134\times10^{-01}$ & 9.$9910\times10^{-03}$ & 1.$8821\times10^{-01}$ & 2.$6034\times10^{-02}$ \\
            Finetuning     & Prop.0.99 & 4.$0291\times10^{-02}$ & 6.$3477\times10^{-04}$ & 8.$2630\times10^{-01}$ & 1.$0961\times10^{-02}$ & 1.$2827\times10^{00}$             & 6.$9390\times10^{-02}$ \\
            Finetuning     & Prop.0.95 & 3.$3487\times10^{-02}$ & 3.$8573\times10^{-04}$ & 8.$4152\times10^{-01}$ & 1.$0287\times10^{-02}$ & 7.$4576\times10^{-01}$ & 5.$7952\times10^{-02}$ \\
            Finetuning     & Mag.0.99  & 3.$3794\times10^{-02}$ & 4.$0906\times10^{-04}$ & 8.$4274\times10^{-01}$ & 1.$0188\times10^{-02}$ & 7.$9177\times10^{-01}$ & 5.$8910\times10^{-02}$ \\
            Finetuning     & Mag.0.95  & 3.$1548\times10^{-02}$ & 3.$3275\times10^{-04}$ & 8.$4295\times10^{-01}$ & 1.$0261\times10^{-02}$ & 4.$3821\times10^{-01}$ & 4.$4580\times10^{-02}$ \\
            \midrule
            Updating       & All       & 1.$2471\times10^{-02}$ & 1.$8916\times10^{-04}$ & 9.$2509\times10^{-01}$ & 6.$1500\times10^{-03}$ & 2.$2029\times10^{-01}$ & 3.$4492\times10^{-04}$ \\
            Updating       & Prop.0.99 & 1.$2471\times10^{-02}$ & 1.$8916\times10^{-04}$ & 9.$2509\times10^{-01}$ & 6.$1500\times10^{-03}$ & 2.$1979\times10^{-01}$ & 3.$4031\times10^{-04}$ \\
            Updating       & Prop.0.95 & 1.$2471\times10^{-02}$ & 1.$8916\times10^{-04}$ & 9.$2509\times10^{-01}$ & 6.$1500\times10^{-03}$ & 2.$1953\times10^{-01}$ & 2.$7152\times10^{-04}$ \\
            Updating       & Mag.0.99  & 3.$2104\times10^{-02}$ & 6.$1611\times10^{-04}$ & 8.$7755\times10^{-01}$ & 7.$7092\times10^{-03}$ & 2.$1284\times10^{-01}$ & 3.$3207\times10^{-04}$ \\
            Updating       & Mag.0.95  & 2.$1411\times10^{-02}$ & 3.$4306\times10^{-04}$ & 8.$9254\times10^{-01}$ & 7.$5031\times10^{-03}$ & 2.$1322\times10^{-01}$ & 3.$3331\times10^{-04}$ \\
            \bottomrule
        \end{tabular}
    \end{table}

    \subsection{Simulated Type-II Diabetic PAtient with Insulin Sensitivity Drift}\label{subsec:glucose_details}

    \begin{figure}
        \centerline{\includegraphics[width=183mm]{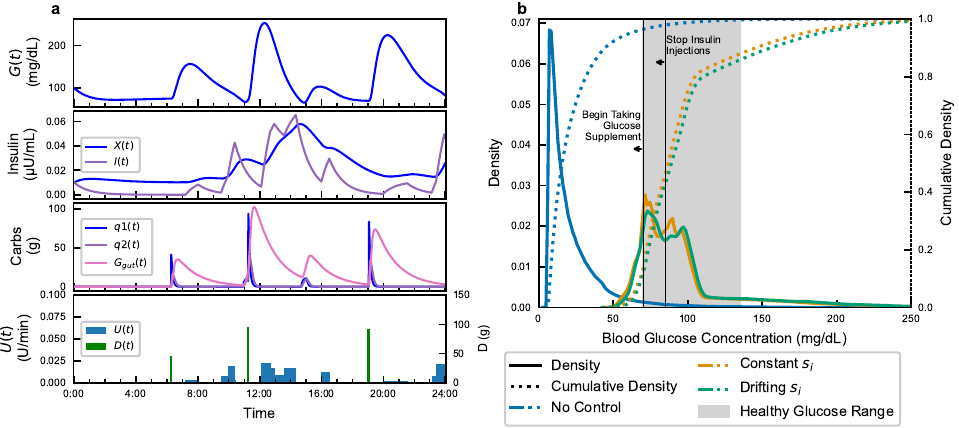}}
        \phantomsubfigure{fig:glucose_exampleday}
        \phantomsubfigure{fig:glucose_distribution}
        \caption{\textbf{a} shows the dynamics between glucose ingestion, delivered insulin, and glood glucose levels in a type II diabetic patient. The top 3 graphs are the predicted states of the system $\bfx$ while the inputs $\bfu$ consist of the initial conditions of the states, and the exogenous inputs $U(t)$ and $D(t)$. \textbf{b} shows rudimentary control scheme for blood glucose control that replicates a patient's manual control. Without this control scheme, blood glucose levels would be almost-constantly below the hypoglycemic level.}
    \end{figure}

    \textbf{Model Details:} Diabetic patients manage blood glucose levels by increasing insulin concentrations to lower blood sugar levels, and by consuming carbohydrates to raise blood sugar levels. In order to generate representative data, we simulate 5 years of data with a simple control system to mantain realistic blood glucose levels: $U(t)$ is generated from a random walk within acceptable bounds, however if blood glucose levels are below 85 mg/dL, $U(t)$ ramps down to 0 for the next 45 minutes. In addition, if blood glucose levels drop below 70 mg/dL, $D(t)$ is increased by 2 g. Parameters used in the simulation are shown in \Cref{tab:glucose_parameters}.
    Meals were simulated with uniform probability between mean begining and end times with a size drawn from a normal distribution with mean and standard deviation shown in the \Cref{tab:glucose_parameters}.

    \begin{table}
        \caption{Parameters for the Diabetic Glucose Model}
        \label{tab:glucose_parameters}
        \begin{tabular}{lrr}
            \toprule
            Parameter              & Value                 & Units                        \\
            \midrule
            $p_1$                  & 1.57$\times$10$^{-2}$ & min$^{-1}$                   \\
            $G_b$                  & 100                   & mg/dL                        \\
            $s_i$                  & 5.0$\times$10$^{-1}$  & L/(min$\cdot$U)              \\
            $\frac{fk_{abs}}{V_G}$ & 8.0$\times$10$^{-2}$  & mg/(dL$\cdot$min$\cdot$gCHO) \\
            $p_2$                  & 1.23$\times$10$^{-2}$ & min$^{-1}$                   \\
            $I_b$                  & 4.0$\times$10$^{-2}$  & U/L                          \\
            $k_e$                  & 1.82$\times$10$^{-2}$ & min$^{-1}$                   \\
            $k_{emp}$              & 1.8$\times$10$^{-1}$  & min$^{-1}$                   \\
            $k_{abs}$              & 1.2$\times$10$^{-2}$  & min$^{-1}$                   \\
            $D(t)$                 & [0, 200]              & gCHO                         \\
            $U(t)$                 & [0, 0.1]              & U/min                        \\
            \midrule
            Breakfast time         & 6:00 AM - 9:00 AM                                    \\
            Breakfast size         & $\mathcal{N}(60,20)$  & gCHO                         \\
            Lunch time             & 11:00 AM - 1:00 PM                                   \\
            Lunch size             & $\mathcal{N}(90,30)$  & gCHO                         \\
            Dinner time            & 5:00 PM - 8:00 PM                                    \\
            Dinner size            & $\mathcal{N}(90,30)$  & gCHO                         \\
            \bottomrule
        \end{tabular}
    \end{table}

    \textbf{Model Architecture:} NODE with 3 states, 1 exogenous input, 2 hidden layers with Sigmoid nonlinearity and 32 neurons each.

    \textbf{Training procedure:} Five years of simulated data were used for model training and validation. Data was normalized before training/inference. The Adam optimizer was used to train the model with the best model selected from the result of the grid-search of hyperparameters shown below in table \Cref{tab:glucose_hyperparameters}.
    \begin{table}
        \caption{Hyperparameters for Diabetic Glucose Model}
        \label{tab:glucose_hyperparameters}
        \begin{tabular}{ll}
            \toprule
            Hyperparameter  & Values                                                                                                     \\
            \midrule
            Learning Rate   & $1\times10^{-4}$, $1\times10^{-3}$, $5\times10^{-3}$, $1\times10^{-2}$. $5\times10^{-2}$. $1\times10^{-1}$ \\
            Batch Size      & $2^6$, $2^7$, $2^8$ $2^9$, $2^{10}$, $2^{11}$, $2^{12}$, $2^{13}$                                          \\
            Layer Sizes     & 32, 64, 128                                                                                                \\
            Training Epochs & 150, 300                                                                                                   \\
            \bottomrule
        \end{tabular}
    \end{table}

    \textbf{Model Maintenance:} Models were trained over the one-year maintenance period with results shown in \Cref{tab:glucose_results}. When finetuning, the loss threshold was 0.06 MSE, learning rate was $1\times10^{-5}$, 50 finetuning epochs were used with batch size 5 and a moving horizon of the last 50 minutes.

    \begin{table}
        \caption{Diabetic Patient Glucose Results}
        \label{tab:glucose_results}
        \begin{tabular}{llrrrrrr}
            \toprule
            Method         & Parameters & MSE                    & MSE 95\% CI            & FSS                     & FSS 95\% CI            & Time                   & Time 95\% CI           \\
            \midrule
            Training       & -          & 6.$2565\times10^{-02}$ & 2.$8609\times10^{-03}$ & -                       & -                      & -                      & -                      \\
            Validation     & -          & 6.$1380\times10^{-02}$ & 7.$1034\times10^{-03}$ & -                       & -                      & -                      & -                      \\
            Test           & -          & 6.$6601\times10^{-02}$ & 8.$7966\times10^{-03}$ & -                       & -                      & -                      & -                      \\
            \midrule
            No Maintenance & -          & 6.$0351\times10^{-02}$ & 5.$6372\times10^{-03}$ & -                       & -                      & -                      & -                      \\
            \midrule
            Finetuning     & full       & 6.$3737\times10^{-02}$ & 4.$4635\times10^{-03}$ & -5.$6112\times10^{-02}$ & 9.$6778\times10^{-02}$ & 5.$3556\times10^{-01}$ & 1.$1063\times10^{-01}$ \\
            Finetuning     & Prop.95    & 6.$2661\times10^{-02}$ & 5.$4937\times10^{-03}$ & -3.$8275\times10^{-02}$ & 1.$8865\times10^{-02}$ & 7.$4542\times10^{-01}$ & 1.$3684\times10^{-01}$ \\
            Finetuning     & Prop.99    & 6.$2142\times10^{-02}$ & 5.$6122\times10^{-03}$ & -2.$9678\times10^{-02}$ & 9.$4840\times10^{-03}$ & 7.$7367\times10^{-01}$ & 1.$3424\times10^{-01}$ \\
            Finetuning     & Mag.95     & 6.$4971\times10^{-02}$ & 4.$6497\times10^{-03}$ & -7.$6553\times10^{-02}$ & 8.$3639\times10^{-02}$ & 5.$8090\times10^{-01}$ & 1.$1955\times10^{-01}$ \\
            Finetuning     & Mag.99     & 6.$3909\times10^{-02}$ & 5.$0067\times10^{-03}$ & -5.$8967\times10^{-02}$ & 4.$1115\times10^{-02}$ & 6.$4684\times10^{-01}$ & 1.$2568\times10^{-01}$ \\
            \midrule
            Updating       & full       & 1.$2049\times10^{-02}$ & 1.$2688\times10^{-03}$ & 8.$0035\times10^{-01}$  & 1.$3260\times10^{-02}$ & 4.$6534\times10^{-01}$ & 3.$7505\times10^{-04}$ \\
            Updating       & Prop.95    & 3.$2093\times10^{-02}$ & 2.$7876\times10^{-03}$ & 4.$6823\times10^{-01}$  & 1.$6340\times10^{-02}$ & 2.$8586\times10^{-01}$ & 2.$2273\times10^{-04}$ \\
            Updating       & Prop.99    & 4.$9722\times10^{-02}$ & 4.$2865\times10^{-03}$ & 1.$7612\times10^{-01}$  & 2.$0324\times10^{-02}$ & 2.$8291\times10^{-01}$ & 2.$1140\times10^{-04}$ \\
            Updating       & Mag.95     & 3.$5986\times10^{-02}$ & 2.$3295\times10^{-03}$ & 4.$0372\times10^{-01}$  & 4.$0991\times10^{-02}$ & 2.$8422\times10^{-01}$ & 2.$3725\times10^{-04}$ \\
            Updating       & Mag.99     & 5.$2191\times10^{-02}$ & 3.$7178\times10^{-03}$ & 1.$3520\times10^{-01}$  & 5.$1046\times10^{-02}$ & 2.$8538\times10^{-01}$ & 2.$2483\times10^{-04}$ \\
            \bottomrule
        \end{tabular}
    \end{table}

    \subsection{Fluid Catalytic Cracker Unit with Simulated Faults}\label{subsec:FCC_details}

    \begin{figure}[ht]
        \centering
        \centerline{\includegraphics[width=183mm]{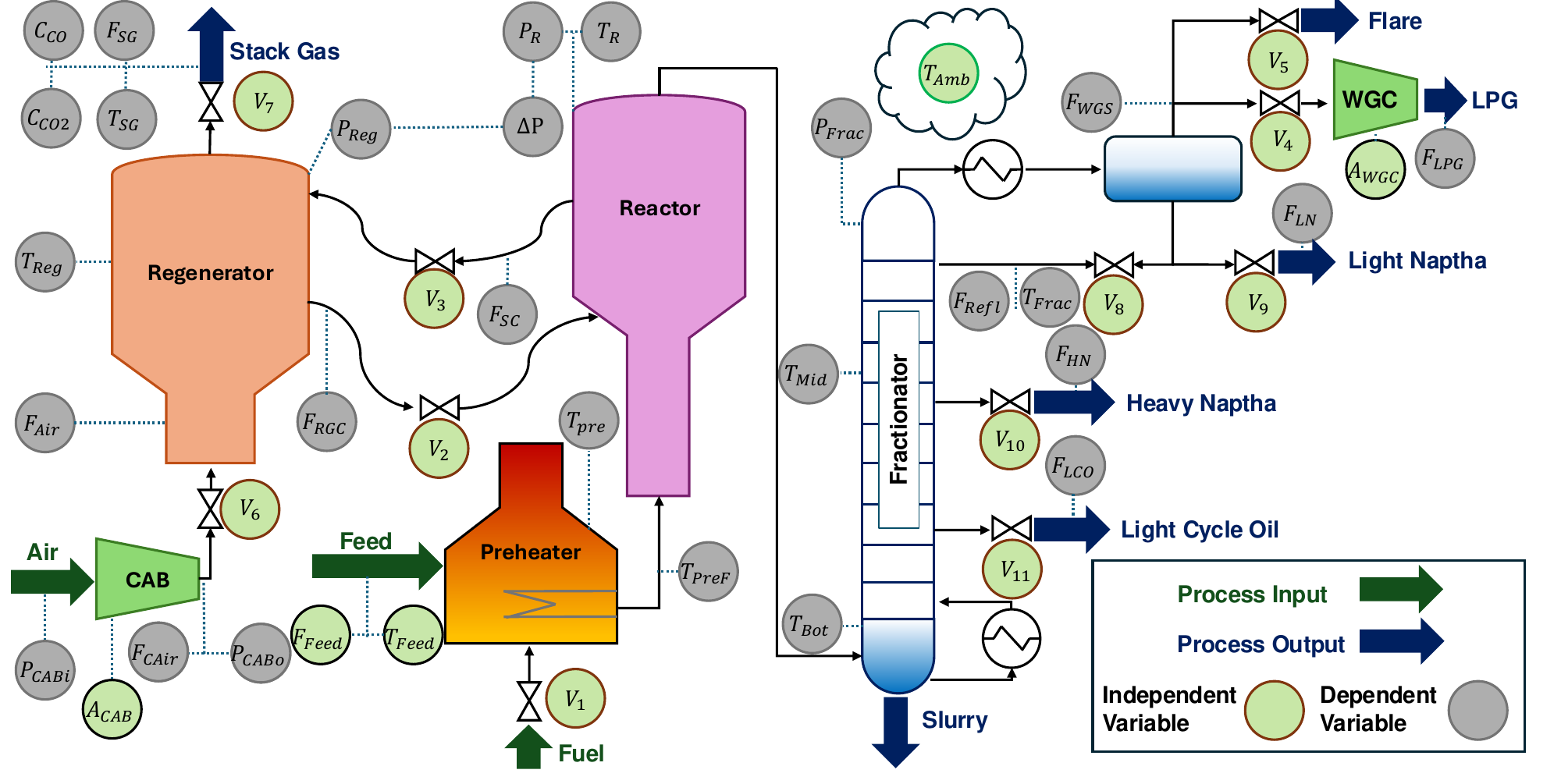}}
        \caption{FCC System\label{fig:FCC_variables}}
    \end{figure}

    \textbf{Model Architecture:} NODE with 31 states, 15 exogenous inputs, 2 hidden layers with Sigmoid nonlinearity and 64 neurons each.

    \textbf{Training procedure:} The FCC-Fractionator model was trained on two datasets: one with normal operating conditions and one with various faults. The normal operating conditions dataset was used for training and validation, while the fault dataset was used for testing. Both vanilla ANNs and NODES performed similarly during training, so NODES were selected with the thought that they would generalize better.

    \begin{table}
        \caption{Variables in the FCC-Fractionator System}
        \label{tab:fcc_variables}
        \begin{tabular}{lll}
            \toprule
            Abbreviation       & Description                     & Units      \\
            \midrule
            31                 & Predicted Variables             & {}         \\
            \midrule
            $C_{CO}$           & Concentration Carbon Monoxide   & ppm        \\
            $C_{CO_2}$         & Concentration Carbon Dioxide    & mol\%      \\
            $F_{Air}$          & Flowrate Air                    & mol/s      \\
            $F_{CAir}$         & Flowrate Combustion Air         & lb/min     \\
            $F_{Fuel}$         & Flowrate Fuel                   & scf/min    \\
            $F_{HN}$           & Flowrate Heavy Naptha           & lb/min     \\
            $F_{LN}$           & Flowrate Light Naptha           & lb/min     \\
            $F_{LCO}$          & Flowrate Light Cycle Oil        & lb/min     \\
            $F_{LPG}$          & Flowrate LPG                    & lb/min     \\
            $F_{RGC}$          & Flowrate Regenerated Catalyst   & lb/min     \\
            $F_{Refl}$         & Flowrate Reflux                 & lb/min     \\
            $F_S$              & Flowrate Slurry                 & lb/min     \\
            $F_{SC}$           & Flowrate Spent Catalyst         & lb/min     \\
            $F_{SG}$           & Flowrate Stack Gas              & mol/min    \\
            $F_{WGC}$          & Flowrate Wet Gas Compressor     & mol/min    \\
            $L_{SP}$           & Level of Stand Pipe             & ft         \\
            $\Delta P$         & Pressure Drop                   & psig       \\
            $P_{CABi}$         & Pressure Combustion Air Inlet   & psia       \\
            $P_{CABo}$         & Pressure Combustion Air Outlet  & psia       \\
            $P_{Frac}$         & Pressure Fractionator           & psia       \\
            $P_{R}$            & Pressure Reactor                & psia       \\
            $P_{Reg}$          & Pressure Regenerator            & psia       \\
            $\Delta T$         & Temperature Difference          & $^\circ$ F \\
            $T_{Bot}$          & Temperature Bottom Fractionator & $^\circ$ F \\
            $T_{Frac}$         & Temperature Fractionator        & $^\circ$ F \\
            $T_{Mid}$          & Temperature Middle Fractionator & $^\circ$ F \\
            $T_{Pre}$          & Temperature Preheater           & $^\circ$ F \\
            $T_{\text{Pre F}}$ & Temperature Preheated Feed      & $^\circ$ F \\
            $T_{Reg}$          & Temperature Regenerator         & $^\circ$ F \\
            $T_{Riser}$        & Temperature Riser               & $^\circ$ F \\
            $T_{SG}$           & Temperature Stack Gas           & $^\circ$ F \\
            \midrule
            15                 & Exogenous Inputs                & {}         \\
            \midrule
            $A_{CAB}$          & Amperage Combustion Air         & Amp        \\
            $A_{WGC}$          & Amperage Wet Gas Compressor     & Amp        \\
            $F_{Feed}$         & Feed Flowrate                   & lb/s       \\
            $T_{Amb}$          & Ambient Temperature             & $^\circ$ F \\
            $T_{Feed}$         & Temperature Feed                & $^\circ$ F \\
            $V_1$              & Valve 1  Position               & \%         \\
            $V_2$              & Valve 2  Position               & \%         \\
            $V_3$              & Valve 3  Position               & \%         \\
            $V_4$              & Valve 4  Position               & \%         \\
            $V_5$              & Valve 5  Position               & \%         \\
            $V_6$              & Valve 6  Position               & \%         \\
            $V_7$              & Valve 7  Position               & \%         \\
            $V_8$              & Valve 8  Position               & \%         \\
            $V_9$              & Valve 9  Position               & \%         \\
            $V_{10}$           & Valve 10  Position              & \%         \\
            \bottomrule
        \end{tabular}
    \end{table}

    \begin{table}
        \caption{Details of the FCC-Fractionator Datasets\label{tab:fcc_datasets}}
        \begin{tabular}{ll}
            \toprule
            Dataset                                             & Length \\
            \midrule
            Normal Operating Conditions - Steady Feed Flowrate  & 2 days \\
            Normal Operating Conditions - Varying Feed Flowrate & 7 days \\
            \midrule
            Heat Exchanger Fouling                              & 1 days \\
            Decreased Condenser Efficiency                      & 1 days \\
            Heavy Naptha Flow Sensor Drift                      & 1 days \\
            Higher Pressure Drop                                & 1 days \\
            CAB Valve Leak                                      & 1 days \\
            \bottomrule
        \end{tabular}
    \end{table}

    \textbf{Model Maintenance:} The FCC-Fractionator model was trained on normal operating conditions and tested on various faults. The model was maintained using retraining and updating with the SEKF. The SEKF was initialized with learning rate 1e-3, q=0.1, and $P_0=100\cdot I$. Finetuning occured using the Adam optimizer and a learning rate of 1e-3 and 50 retraining epochs. The finetuning dataset was comprised of up to the last 50 minutes of data. The results are shown in \Cref{tab:fcc_results_1} and \Cref{tab:fcc_results_2}.

    \begin{landscape}
        \begin{table}
            \caption{Part 1 of FCC-Fractionator Results}
            \label{tab:fcc_results_1}
            \begin{tabular}{lllrrllll}
                \toprule
                Fault                          & Method         & Params. & MSE                    & 95\% CI                & FSS                    & 95\% CI                & Mean Time (s)          & 95\% CI                \\
                \midrule
                -                              & Training       & -       & $2.2858\times10^{-01}$ & $7.3446\times10^{-02}$ & -                      & -                      & -                      & -                      \\
                -                              & Validation     & -       & $1.9714\times10^{-01}$ & $5.4009\times10^{-03}$ & -                      & -                      & -                      & -                      \\
                \midrule
                CAB Valve Leak                 & No Maintenance & -       & $1.0820\times10^{00}$             & $2.2326\times10^{-02}$ & -                      & -                      & -                      & -                      \\
                CAB Valve Leak                 & Finetuning     & full    & $1.9565\times10^{-01}$ & $6.1422\times10^{-03}$ & $7.2415\times10^{-01}$ & $1.4812\times10^{-02}$ & $1.5875 \times 10^{+00}$             & $1.1517\times10^{-01}$ \\
                CAB Valve Leak                 & Finetuning     & Prop.95 & $2.2494\times10^{-01}$ & $1.1762\times10^{-02}$ & $7.0367\times10^{-01}$ & $1.5655\times10^{-02}$ & $1.5426 \times 10^{+00}$             & $1.1677\times10^{-01}$ \\
                CAB Valve Leak                 & Finetuning     & Prop.99 & $2.3210\times10^{-01}$ & $1.3312\times10^{-02}$ & $6.9916\times10^{-01}$ & $1.5828\times10^{-02}$ & $1.8069 \times 10^{+00}$             & $1.2471\times10^{-01}$ \\
                CAB Valve Leak                 & Finetuning     & Mag.95  & $2.1772\times10^{-01}$ & $9.8060\times10^{-03}$ & $7.0818\times10^{-01}$ & $1.5318\times10^{-02}$ & $2.0561 \times 10^{+00}$             & $1.3556\times10^{-01}$ \\
                CAB Valve Leak                 & Finetuning     & Mag.99  & $2.2022\times10^{-01}$ & $1.0479\times10^{-02}$ & $7.0684\times10^{-01}$ & $1.5424\times10^{-02}$ & $1.8135 \times 10^{+00}$             & $1.3062\times10^{-01}$ \\
                CAB Valve Leak                 & Updating       & full    & $1.9786\times10^{-01}$ & $9.1068\times10^{-03}$ & $7.2533\times10^{-01}$ & $1.5204\times10^{-02}$ & $7.4818\times10^{-01}$ & $1.3389\times10^{-03}$ \\
                CAB Valve Leak                 & Updating       & Prop.95 & $2.0919\times10^{-01}$ & $9.1226\times10^{-03}$ & $7.1843\times10^{-01}$ & $1.4850\times10^{-02}$ & $6.2338\times10^{-01}$ & $4.3430\times10^{-04}$ \\
                CAB Valve Leak                 & Updating       & Prop.99 & $2.1008\times10^{-01}$ & $8.8157\times10^{-03}$ & $7.1693\times10^{-01}$ & $1.4853\times10^{-02}$ & $5.8055\times10^{-01}$ & $1.0839\times10^{-03}$ \\
                CAB Valve Leak                 & Updating       & Mag.95  & $2.7196\times10^{-01}$ & $3.8912\times10^{-02}$ & $6.6303\times10^{-01}$ & $3.3941\times10^{-02}$ & $5.9998\times10^{-01}$ & $1.6162\times10^{-03}$ \\
                CAB Valve Leak                 & Updating       & Mag.99  & $2.5378\times10^{-01}$ & $2.3522\times10^{-02}$ & $6.7813\times10^{-01}$ & $2.3023\times10^{-02}$ & $5.9495\times10^{-01}$ & $1.6473\times10^{-03}$ \\
                \midrule
                Decreased Condenser Efficiency & No Maintenance & -       & $2.2849\times10^{-01}$ & $2.5593\times10^{-03}$ & -                      & -                      & -                      & -                      \\
                Decreased Condenser Efficiency & Finetuning     & full    & $1.7040\times10^{-01}$ & $3.9717\times10^{-04}$ & $2.2173\times10^{-01}$ & $8.3153\times10^{-03}$ & $8.1784\times10^{-01}$ & $8.7084\times10^{-02}$ \\
                Decreased Condenser Efficiency & Finetuning     & Prop.95 & $1.7077\times10^{-01}$ & $4.6047\times10^{-04}$ & $2.2030\times10^{-01}$ & $8.3225\times10^{-03}$ & $9.4395\times10^{-01}$ & $9.0861\times10^{-02}$ \\
                Decreased Condenser Efficiency & Finetuning     & Prop.99 & $1.7112\times10^{-01}$ & $4.6744\times10^{-04}$ & $2.1892\times10^{-01}$ & $8.2880\times10^{-03}$ & $1.0597 \times 10^{+00}$             & $9.9739\times10^{-02}$ \\
                Decreased Condenser Efficiency & Finetuning     & Mag.95  & $1.7097\times10^{-01}$ & $4.6100\times10^{-04}$ & $2.1924\times10^{-01}$ & $8.3764\times10^{-03}$ & $1.1013 \times 10^{+00}$             & $9.5548\times10^{-02}$ \\
                Decreased Condenser Efficiency & Finetuning     & Mag.99  & $1.7109\times10^{-01}$ & $4.7182\times10^{-04}$ & $2.1923\times10^{-01}$ & $8.2496\times10^{-03}$ & $1.0555 \times 10^{+00}$             & $9.8528\times10^{-02}$ \\
                Decreased Condenser Efficiency & Updating       & full    & $1.6499\times10^{-01}$ & $4.7353\times10^{-04}$ & $2.4611\times10^{-01}$ & $8.2568\times10^{-03}$ & $7.0269\times10^{-01}$ & $3.2961\times10^{-04}$ \\
                Decreased Condenser Efficiency & Updating       & Prop.95 & $1.6780\times10^{-01}$ & $4.8018\times10^{-04}$ & $2.3355\times10^{-01}$ & $8.3221\times10^{-03}$ & $5.7447\times10^{-01}$ & $3.5543\times10^{-03}$ \\
                Decreased Condenser Efficiency & Updating       & Prop.99 & $1.6900\times10^{-01}$ & $4.8920\times10^{-04}$ & $2.2847\times10^{-01}$ & $8.2842\times10^{-03}$ & $5.6503\times10^{-01}$ & $4.4142\times10^{-04}$ \\
                Decreased Condenser Efficiency & Updating       & Mag.95  & $1.6813\times10^{-01}$ & $4.8340\times10^{-04}$ & $2.3202\times10^{-01}$ & $8.3462\times10^{-03}$ & $5.6570\times10^{-01}$ & $1.6098\times10^{-03}$ \\
                Decreased Condenser Efficiency & Updating       & Mag.99  & $1.6943\times10^{-01}$ & $5.0287\times10^{-04}$ & $2.2695\times10^{-01}$ & $8.1927\times10^{-03}$ & $5.6524\times10^{-01}$ & $4.8120\times10^{-04}$ \\
                \midrule
                Heat Exchanger Fouling         & No Maintenance & -       & $7.7872\times10^{-01}$ & $2.7629\times10^{-02}$ & -                      & -                      & -                      & -                      \\
                Heat Exchanger Fouling         & Finetuning     & full    & $1.7611\times10^{-01}$ & $7.1871\times10^{-04}$ & $5.5694\times10^{-01}$ & $1.8109\times10^{-02}$ & $1.3235 \times 10^{+00}$             & $1.0952\times10^{-01}$ \\
                Heat Exchanger Fouling         & Finetuning     & Prop.95 & $1.8121\times10^{-01}$ & $1.0648\times10^{-03}$ & $5.5209\times10^{-01}$ & $1.8026\times10^{-02}$ & $2.0830 \times 10^{+00}$             & $1.2582\times10^{-01}$ \\
                Heat Exchanger Fouling         & Finetuning     & Prop.99 & $1.8531\times10^{-01}$ & $1.2664\times10^{-03}$ & $5.4797\times10^{-01}$ & $1.7944\times10^{-02}$ & $2.8707 \times 10^{+00}$             & $1.3877\times10^{-01}$ \\
                Heat Exchanger Fouling         & Finetuning     & Mag.95  & $1.7954\times10^{-01}$ & $9.8383\times10^{-04}$ & $5.5300\times10^{-01}$ & $1.8112\times10^{-02}$ & $2.0308 \times 10^{+00}$             & $1.3082\times10^{-01}$ \\
                Heat Exchanger Fouling         & Finetuning     & Mag.99  & $1.8055\times10^{-01}$ & $1.0280\times10^{-03}$ & $5.5187\times10^{-01}$ & $1.8098\times10^{-02}$ & $2.2107 \times 10^{+00}$             & $1.3625\times10^{-01}$ \\
                Heat Exchanger Fouling         & Updating       & full    & $1.6833\times10^{-01}$ & $6.8298\times10^{-04}$ & $5.7228\times10^{-01}$ & $1.7829\times10^{-02}$ & $7.3931\times10^{-01}$ & $4.1766\times10^{-03}$ \\
                Heat Exchanger Fouling         & Updating       & Prop.95 & $1.7398\times10^{-01}$ & $7.8964\times10^{-04}$ & $5.6345\times10^{-01}$ & $1.7974\times10^{-02}$ & $5.6607\times10^{-01}$ & $1.0917\times10^{-03}$ \\
                Heat Exchanger Fouling         & Updating       & Prop.99 & $1.7661\times10^{-01}$ & $8.6650\times10^{-04}$ & $5.5959\times10^{-01}$ & $1.7997\times10^{-02}$ & $5.6679\times10^{-01}$ & $5.8452\times10^{-04}$ \\
                Heat Exchanger Fouling         & Updating       & Mag.95  & $1.7282\times10^{-01}$ & $7.5622\times10^{-04}$ & $5.6408\times10^{-01}$ & $1.8046\times10^{-02}$ & $5.6995\times10^{-01}$ & $2.6068\times10^{-03}$ \\
                Heat Exchanger Fouling         & Updating       & Mag.99  & $1.7474\times10^{-01}$ & $7.9104\times10^{-04}$ & $5.6003\times10^{-01}$ & $1.8133\times10^{-02}$ & $5.8346\times10^{-01}$ & $1.1117\times10^{-03}$ \\
                \bottomrule
            \end{tabular}
        \end{table}

        \begin{table}
            \caption{Part 2 of FCC-Fractionator Results}
            \label{tab:fcc_results_2}
            \begin{tabular}{lllrrllll}
                \toprule
                Fault                          & Method         & Params. & MSE                    & 95\% CI                & FSS                    & 95\% CI                & Mean Time (s)          & 95\% CI                \\
                \midrule
                -                              & Training       & -       & $2.2858\times10^{-01}$ & $7.3446\times10^{-02}$ & -                      & -                      & -                      & -                      \\
                -                              & Validation     & -       & $1.9714\times10^{-01}$ & $5.4009\times10^{-03}$ & -                      & -                      & -                      & -                      \\
                \midrule
                Heavy Naptha Flow Sensor Drift & No Maintenance & -       & $1.9995\times10^{-01}$ & $1.5275\times10^{-03}$ & -                      & -                      & -                      & -                      \\
                Heavy Naptha Flow Sensor Drift & retraining     & full    & $1.6933\times10^{-01}$ & $5.2727\times10^{-04}$ & $1.3777\times10^{-01}$ & $6.2148\times10^{-03}$ & $7.9557\times10^{-01}$ & $8.4901\times10^{-02}$ \\
                Heavy Naptha Flow Sensor Drift & retraining     & Prop.95 & $1.7060\times10^{-01}$ & $5.3188\times10^{-04}$ & $1.3233\times10^{-01}$ & $5.8983\times10^{-03}$ & $9.8887\times10^{-01}$ & $9.3178\times10^{-02}$ \\
                Heavy Naptha Flow Sensor Drift & retraining     & Prop.99 & $1.7054\times10^{-01}$ & $5.4508\times10^{-04}$ & $1.3259\times10^{-01}$ & $5.9484\times10^{-03}$ & $1.1264 \times 10^{+00}$             & $1.0328\times10^{-01}$ \\
                Heavy Naptha Flow Sensor Drift & retraining     & Mag.95  & $1.7038\times10^{-01}$ & $5.5995\times10^{-04}$ & $1.3284\times10^{-01}$ & $6.1827\times10^{-03}$ & $1.1203 \times 10^{+00}$             & $1.0191\times10^{-01}$ \\
                Heavy Naptha Flow Sensor Drift & retraining     & Mag.99  & $1.7060\times10^{-01}$ & $5.4396\times10^{-04}$ & $1.3225\times10^{-01}$ & $5.9552\times10^{-03}$ & $1.1190 \times 10^{+00}$             & $1.0239\times10^{-01}$ \\
                Heavy Naptha Flow Sensor Drift & updating       & full    & $1.6485\times10^{-01}$ & $6.0120\times10^{-04}$ & $1.5979\times10^{-01}$ & $6.5344\times10^{-03}$ & $7.0761\times10^{-01}$ & $1.2636\times10^{-03}$ \\
                Heavy Naptha Flow Sensor Drift & updating       & Prop.95 & $1.6737\times10^{-01}$ & $5.9848\times10^{-04}$ & $1.4716\times10^{-01}$ & $6.5351\times10^{-03}$ & $5.7458\times10^{-01}$ & $2.1382\times10^{-03}$ \\
                Heavy Naptha Flow Sensor Drift & updating       & Prop.99 & $1.6819\times10^{-01}$ & $5.9751\times10^{-04}$ & $1.4329\times10^{-01}$ & $6.4515\times10^{-03}$ & $5.7767\times10^{-01}$ & $6.6518\times10^{-04}$ \\
                Heavy Naptha Flow Sensor Drift & updating       & Mag.95  & $1.6760\times10^{-01}$ & $5.9940\times10^{-04}$ & $1.4604\times10^{-01}$ & $6.5318\times10^{-03}$ & $5.7613\times10^{-01}$ & $9.0233\times10^{-04}$ \\
                Heavy Naptha Flow Sensor Drift & updating       & Mag.99  & $1.6826\times10^{-01}$ & $5.9952\times10^{-04}$ & $1.4295\times10^{-01}$ & $6.4578\times10^{-03}$ & $5.7301\times10^{-01}$ & $7.1910\times10^{-04}$ \\
                \midrule
                Higher Pressure Drop           & No Maintenance & -       & $5.8419\times10^{01}$             & $1 \times 10^{+00}$             & -                      & -                      & -                      & -                      \\
                Higher Pressure Drop           & retraining     & full    & $2.9989\times10^{-01}$ & $2.8644\times10^{-02}$ & $9.2004\times10^{-01}$ & $1.3100\times10^{-02}$ & $3.2292$             & $1.4362\times10^{-01}$ \\
                Higher Pressure Drop           & retraining     & Prop.95 & $4.0032\times10^{-01}$ & $4.5861\times10^{-02}$ & $9.1528\times10^{-01}$ & $1.3335\times10^{-02}$ & $4.8977$             & $5.9930\times10^{-02}$ \\
                Higher Pressure Drop           & retraining     & Prop.99 & $4.1864\times10^{-01}$ & $4.5468\times10^{-02}$ & $9.1330\times10^{-01}$ & $1.3453\times10^{-02}$ & $5.4246$             & $6.4408\times10^{-02}$ \\
                Higher Pressure Drop           & retraining     & Mag.95  & $3.7784\times10^{-01}$ & $4.5928\times10^{-02}$ & $9.1663\times10^{-01}$ & $1.3283\times10^{-02}$ & $4.9987$             & $9.2403\times10^{-02}$ \\
                Higher Pressure Drop           & retraining     & Mag.99  & $3.8487\times10^{-01}$ & $4.5973\times10^{-02}$ & $9.1577\times10^{-01}$ & $1.3388\times10^{-02}$ & $5.0694$             & $7.0485\times10^{-02}$ \\
                Higher Pressure Drop           & updating       & full    & $4.4462\times10^{-01}$ & $4.3729\times10^{-02}$ & $9.2038\times10^{-01}$ & $1.2747\times10^{-02}$ & $7.6581\times10^{-01}$ & $1.4917\times10^{-03}$ \\
                Higher Pressure Drop           & updating       & Prop.95 & $3.4164\times10^{-01}$ & $2.3384\times10^{-02}$ & $9.2006\times10^{-01}$ & $1.3014\times10^{-02}$ & $6.2386\times10^{-01}$ & $8.4890\times10^{-04}$ \\
                Higher Pressure Drop           & updating       & Prop.99 & $3.5197\times10^{-01}$ & $2.3412\times10^{-02}$ & $9.1898\times10^{-01}$ & $1.3127\times10^{-02}$ & $7.8509\times10^{-01}$ & $2.9385\times10^{-02}$ \\
                Higher Pressure Drop           & updating       & Mag.95  & $3.9371\times10^{-01}$ & $2.7744\times10^{-02}$ & $9.1953\times10^{-01}$ & $1.3008\times10^{-02}$ & $6.3863\times10^{-01}$ & $1.8718\times10^{-03}$ \\
                Higher Pressure Drop           & updating       & Mag.99  & $3.9231\times10^{-01}$ & $3.2431\times10^{-02}$ & $9.1845\times10^{-01}$ & $1.3184\times10^{-02}$ & $6.3380\times10^{-01}$ & $1.1499\times10^{-03}$ \\
                \bottomrule
            \end{tabular}
        \end{table}
    \end{landscape}
\end{appendices}


\end{document}